\documentclass[11pt]{article}

\usepackage[margin=1.in]{geometry}

\usepackage{caption}
\usepackage{url}
\usepackage{wrapfig}
\usepackage[T1]{fontenc}
\usepackage[utf8]{inputenc}
\usepackage{authblk}
\usepackage{color,xcolor}
\usepackage{amsmath, amssymb, mathtools, latexsym, natbib}
\usepackage{psfrag,epsfig,amsfonts,latexsym,amsthm,,amscd,url }
\usepackage{grffile}
\usepackage{bm}
\usepackage{appendix}
\usepackage{subcaption}
\usepackage{booktabs} 
\usepackage[version=4]{mhchem}
\usepackage[colorlinks=true,linkcolor=blue,citecolor=blue, linktocpage=true]{hyperref}

\allowdisplaybreaks[4]

\newcommand{\eps}{\varepsilon}

\renewcommand{\phi}{\varphi}

\newcommand{\dd}{{\rm d}}
\newcommand{\x}{{\bm x}}
\newcommand{\X}{{\bm X}}

\theoremstyle{plain}
\newtheorem{theorem}{Theorem}[section]

\theoremstyle{definition}

\theoremstyle{remark}
\newtheorem{remark}[theorem]{Remark}
\newtheorem{example}[theorem]{Example}

\newcommand{\BE}{\begin{equation}}
\newcommand{\EE}{\end{equation}}
\newcommand{\BEN}{\begin{equation*}}
\newcommand{\EEN}{\end{equation*}}
\newcommand{\BAL}{\begin{align}}
\newcommand{\EAL}{\end{align}}
\newcommand{\BAN}{\begin{align*}}

\DeclareMathOperator*{\argmin}{arg\,min}

\begin{document}

\title{Structure-Aware Variational Learning of a Class of Generalized Diffusions}

\author[1,2]{Yubin Lu\thanks{Corresponding author: ylu117@scut.edu.cn}}
\author[2]{Xiaofan Li}
\author[2]{Chun Liu}
\author[3]{Qi Tang}
\author[4]{Yiwei Wang}
\affil[1]{Current Address: School of Mathematics, South China University of Technology, Guangzhou 510640, China.}
\affil[2]{Department of Applied Mathematics, Illinois Institute of Technology, Chicago, IL 60616 United States}
\affil[3]{School of Computational Science and Engineering, Georgia Institute of Technology, Atlanta, GA 30332 United States}
\affil[4]{Department of Mathematics University of California Riverside, Riverside, CA 92521 United States}

\renewcommand*{\Affilfont}{\small\it} 
\renewcommand\Authands{ and } 
\date{\today}
\maketitle

\begin{abstract}
Learning the underlying potential energy of gradient systems from partial and noisy observations is a fundamental problem arising in physics, chemistry, and data-driven modeling. Classical approaches often rely on direct regression of governing equations or velocity fields, which can be sensitive to noise and external perturbations and may fail when observations are incomplete. In this work, we propose a structure-aware, variational learning framework for inferring unknown potential functions in generalized diffusions, grounded in the energetic variational approach. Starting from the energy-dissipation law associated with the Fokker-Planck equation, we construct loss functions based on the De Giorgi-type energy-dissipation functional, which consistently couples the free energy and the dissipation mechanism of the system. This formulation avoids explicit enforcement of the governing partial differential equation and preserves the underlying variational structure of the dynamics. Through numerical experiments in one, two, and three dimensions, we demonstrate that the proposed energy-dissipation-law-based loss exhibits enhanced robustness with respect to observation time, noise level, and the diversity and amount of available training data. These results highlight the effectiveness of energy-dissipation principles as a reliable foundation for learning stochastic diffusion dynamics from data.
\end{abstract}

{\small \medskip\par\noindent
{\bf Key Words}: Energetic variational approach; Data-driven modeling; Fokker-Planck equation; De Giorgi-type energy-dissipation functional.}

\section{Introduction}\label{sec:intro}
\indent Variational models play an important role in studying a wide range of problems in physics, materials science, biology, and data science
\cite{doi2011onsager, weinan2020machine, peletier2014variational}. A defining feature of these models is that they describe how complex systems evolve through their energy and their irreversible loss of energy over time via an energy-dissipation law:
$$
\frac{\mathrm{d}}{\mathrm{d} t} E({\bm x}) = - \triangle({\bm x}, \dot{\bm x}).
$$
Here ${\bm x}$ denotes the state variables of interest (e.g., fluid density, order parameters, or concentrations), $\dot{\bm x}$ is their rate of change, $E=\mathcal F+\mathcal K$ is the total energy consisting of free energy $\mathcal F$ and kinetic energy $\mathcal K$, and $\triangle$ is the dissipation rate. The energy part captures multiscale couplings and competitions in the system, while the dissipation part encodes how the system relaxes toward equilibrium. In isothermal and mechanically closed systems, the existence of an energy–dissipation law is a direct consequence of the first and second laws of thermodynamics \cite{ericksen1992introduction}, and thus provides a universal and physically interpretable organizing principle.

Over the past few decades, variational modeling frameworks have proven particularly powerful for complex fluids \cite{doi2011onsager, EnVarA1, grmela1997dynamics, ottinger1997dynamics, peletier2014variational, wang2022some, wu2007density}, i.e., fluids with internal structures such as orientational order of rod-like molecules, elasticity of deformable particles, and electro-chemo-mechanical couplings induced by ions and polymers \cite{doi2011onsager, liu2009introduction, grmela1997dynamics}. Compared with traditional force-by-force modeling, variational approaches offer a systematic route to incorporate multiscale chemo-mechanical effects and, crucially, to enforce thermodynamic consistency by construction. Classically, the development of a variational model is \emph{mechanism-driven}: one starts from physical principles and modeling assumptions (e.g., constitutive laws, symmetry, and coarse-graining arguments), postulates functional forms for the free energy and dissipation mechanisms, and then derives governing equations or evolution laws.  This workflow has produced many successful models, but it faces well-known bottlenecks in complex multiscale systems. Key energetic or dissipative ingredients may be unknown, difficult to parameterize, or expensive to calibrate, and distinct modeling choices can yield similar qualitative behavior but markedly different quantitative predictions. These challenges have motivated a growing \emph{data-driven} paradigm in which observational or simulation data are used to inform, calibrate, or even discover components of the variational structure \cite{weinan2020machine, yang2026data}.

Learning physical laws from data is now one of the central themes in scientific machine learning. One of important examples is learning generalized diffusion processes from appropriately observed data, which has been widely studied across various scientific and engineering disciplines. Numerous approaches have been proposed to infer unknown quantities in generalized diffusion models, including probabilistic/statistical inference methods \cite{Felix, VI4SDE, LiuSWasserstein, GP4PDEs1, GP4PDEs2, Kernel4PDEs}, physics-informed neural networks in strong formulations \cite{PINN, PINNinverse1, PINNinverse2}, and weak formulations \cite{weakSINDy, weakSINDy4PDE, Lu_self-test, DGM, neuralGalerkin, weakPINNs, variationalPINN, weakSINDy4meanfield, Phase-field_DeepONet, EntropyInformed, zhang2022gfinns, huang2022variational, MLforIrreversibleProcess1, MLforIrreversibleProcess2, lu2025EnVarA}, to cite a few. However, most of these methods are formulated at the level of the governing PDE/SDE itself and rely, either explicitly or implicitly, on enforcing local (in time and space) equation residuals or moment identities, which can be sensitive to derivative estimation, discretization choices, and model mismatch when data are noisy or sparsely sampled.

In this work, we propose a new learning paradigm that starts from the \emph{variational structure} of irreversible dynamics. Our goal is to design learning algorithms directly from the energy–dissipation law, thereby refining unknown components of the variational model while maintaining thermodynamic consistency. As a prototypical toy problem, we consider overdamped Langevin dynamics with friction $\gamma>0$,
\begin{equation}\label{eq:sde}
dX_t = -\frac{1}{\gamma}\nabla \psi(X_t) \, dt + \sqrt{\frac{2 k_B T}{\gamma}} \, dW_t,
\end{equation}
where $\psi(x)$ is an unknown potential. The associated density $\rho(\bm x,t)$ satisfies a Fokker–Planck equation 
\begin{equation}\label{eq:fpe}
 \rho_t  = \nabla \cdot \left( \frac{1}{\gamma}  \rho \nabla (k_B T \ln \rho + \psi ) \right)
\end{equation}
which admits an energy-dissipation law
\begin{equation}\label{ED_1}
 \frac{\dd}{\dd t} \int_{\Omega} k_B T \rho \ln \rho + \rho \psi  \dd x = - \int \gamma \rho |{\bm v}|^2 \dd x
\end{equation}
along with the mass conservation $\rho_t  + \nabla \cdot (\rho {\bm v} ) = 0$ and ${\bm v}=-\frac{1}{\gamma}\nabla(K_BT\ln\rho+\psi)$. In what follows, we set $\gamma=1$ for simplicity, as our focus is on learning the potential $\psi$.

In the paradigm of variational modeling, we assume that the available data consist of particle trajectory snapshots generated by the SDE (\ref{eq:sde}) at a few discrete times (possibly from multiple ensembles with different initial conditions). The goal is to recover the external potential $\psi$ in (\ref{ED_1}). From a modeling perspective, we can view the data as coming from the miscroscopic stochastic simulation, while our goal is to learn a coarse-grained Fokker–Planck description in terms of density $\rho$. In this sense, our setting can be viewed as a simple yet informative instance of data-driven \emph{coarse-graining}, where we infer the effective variational structure governing the evolution of an empirical density from stochastic particle data. A major conceptual difficulty is that an energy--dissipation law, while an excellent \emph{modeling} principle, does not automatically translate into a practical \emph{learning} objective. By itself, an energy--dissipation law is not an optimization principle: it states that along a \emph{true} trajectory the total energy decreases at a rate prescribed by the dissipation, but it does not directly provide an objective whose minimizer identifies the unknown model components. In particular, \eqref{ED_1} is an equality  constraint on \(({\bm x}(t),\dot{\bm x}(t)) \), rather than a variational characterization that determines \({\bm x}(t)\) through a minimization problem at each time.

In our recent work \cite{lu2025EnVarA}, we took a first step toward bridging this gap by constructing a learning algorithm based on a discretized \emph{differential-form} energy balance. The numerical results in \cite{lu2025EnVarA} demonstrate improved robustness to corrupted observations compared with PDE-residual-based losses. At the same time, \cite{lu2025EnVarA} also reveals that the resulting loss may suffer from nontrivial \emph{identifiability} issues and can fail in certain regimes. Beyond these challenges, there is a more subtle yet fundamental issue that has received relatively little attention: even when different formulations of a learning objective are equivalent at the continuous level, their discrete counterparts, as implemented in practice, can behave quite differently. In particular, many learning principles derived from variational structures, such as those based on PDEs and variational principles, are formally equivalent when evaluated along sufficiently smooth solutions. However, once discretized in time and space, these formulations may respond very differently to sampling error, noise, and model misspecification. As a result, losses that are theoretically interchangeable can lead to significantly different optimization landscapes, stability properties, and ultimately, learning outcomes. This observation suggests that the choice of loss functional in data-driven variational modeling should not be guided solely by its continuous formulation, but must also take into account its discrete realization and numerical robustness. In this work, we revisit the design of learning objectives from this perspective and propose a formulation based on a De~Giorgi-type energy--dissipation functional over finite time intervals \cite{deGiorgi1,deGiorgi2,ambrosio2005gradient}, which naturally incorporates temporal coarse-graining and leads to improved stability and identifiability in practice.

A closely related issue lies in the \emph{level of observables} used to formulate the learning problem. In practice, available data often consist of discrete particle trajectories generated by an underlying stochastic dynamics, while quantities such as velocity fields or fluxes must be inferred indirectly from these observations. This mismatch introduces an additional layer of approximation, and in many cases leads to systematic biases, especially under sparse temporal sampling or measurement noise. When the learning objective is formulated at the level of such derived quantities, these biases are inevitably propagated into the inferred model, potentially leading to incorrect identification of the underlying energetics. In this sense, the learning problem can become effectively non-identifiable under imperfect observations, even if the underlying dynamics are well-defined. By contrast, we first emphasize that velocity information, even when obtained in a coarse or significantly biased form, can still provide valuable guidance for learning the underlying dynamics. Such information may encode useful local directional structure of the evolution and can be beneficial when properly incorporated. However, directly coupling this velocity information with the learning objective may lead to instability, as any systematic bias in the observed velocities is directly injected into the inferred model. To address this issue, the proposed method instead achieves a structural separation between the role of velocity observations and the core learning problem. In particular, the learning framework is formulated so that velocity information does not enter as a direct matching target, but is instead implicitly incorporated through the variational energy--dissipation structure over finite time intervals. This separation leads to a more stable learning formulation that is less sensitive to biased or incomplete velocity observations, while still fully utilizing the information contained in trajectory data to recover the underlying potential.

The remainder of this paper is organized as follows. In Section 2, we review the background on generalized diffusion processes and the variational framework that underlies our approach. Section 3 introduces the proposed structure-aware learning methodology, highlighting how different modeling components are integrated within the variational formulation. In Section 4, we present numerical experiments that demonstrate the effectiveness of the approach under various data conditions. Finally, Section 5 concludes the paper with discussions on potential extensions and future research directions.


\section{Preliminary}
This section begins with an introduction to the Energetic Variational Approach (EnVarA), which forms the core theoretical framework of this study. We then provide a brief review of related methods and discuss their relevance to the present work.

\subsection{An Energetic Variational Approach (EnVarA)}\label{sec:EnVarA}
We briefly recall the energetic variational structure and the associated variational formulations, which will be used later to motivate the design of loss functions.

Inspired by nonequilibrium thermodynamics, particularly the pioneering works of Rayleigh \cite{Rayleigh} and Onsager \cite{Onsager1,Onsager2}, the EnVarA has become a powerful framework for investigating a broad class of complex fluid systems in physics \cite{eisenberg2022variational}, chemistry \cite{wang2020field}, and biochemistry \cite{wang2021two}.

To convey the core idea without loss of generality, we begin with a simple yet representative setting. Specifically, we consider an isothermal, mechanically isolated complex system whose dynamics obey an energy–dissipation law of the form
\begin{equation}\label{eqn:original energylaw}
    \frac{d}{dt}E=-\Delta\leq 0,
\end{equation}
where the total energy $E$ consists of the kinetic energy $\mathcal{K}$ and the Helmholtz free energy $\mathcal{F}$, and $\Delta$ denotes the instantaneous rate of energy dissipation. 

EnVarA offers a systematic and thermodynamically consistent framework to derive the governing equations of a dynamical system based on the energy–dissipation law \eqref{eqn:original energylaw}, by combining the least action principle (LAP) and the maximum dissipation principle (MDP).

Specifically, the conservative (Hamiltonian) part of the dynamics is obtained by applying the LAP to the action functional
$$
\mathcal{A}({\bm x}) = \int_0^T \left( \mathcal{K} - \mathcal{F} \right) \dd t,
$$
where $\bm x$ denotes the trajectory in Lagrangian coordinates. Taking the variation of $\mathcal{A}$ with respect to $\bm x$ yields
 $$
 \delta \mathcal{A} =  \int_{0}^T \int_{\Omega} ({\rm force}_{\text{iner}} - {\rm force}_{\text{conv}})\cdot \delta {\bm x}  ~ \dd \x \dd t,
 $$
where $\Omega$ denotes the spatial domain, and ${\rm force}_{\text{iner}}$ and ${\rm force}_{\text{conv}}$ represent the inertial and conservative forces, respectively \cite{EnVarA1, arnol2013mathematical}.

The dissipative part of the dynamics is derived from the MDP by taking the variation of the Onsager dissipation functional $\mathcal{D}$ with respect to the rate $\dot{\bm x}$. In the linear response regime \cite{Onsager2}, the dissipation functional is given by $\mathcal{D} = \frac{1}{2}\Delta$, and its variation yields
$$
\delta \mathcal{D}  = \int_{\Omega} {\rm force}_{\text{diss}} \cdot \delta \dot{\x}~ \dd \x,
$$
where ${\rm force}_{\text{diss}}$ denotes the dissipative force.

The evolution equation of the system is then obtained by enforcing the force balance condition between the conservative and dissipative forces, namely,
\begin{equation}
\frac{\delta \mathcal{D}}{\delta \dot{\bm x}}=\frac{\delta \mathcal{A}}{\delta \bm x}\nonumber.
\end{equation}
The detailed variational derivations leading to these expressions are standard and can be found in \cite{EnVarA1}.
\paragraph{\bf Energetic Variational Formulation of the Fokker--Planck Equation}
As a canonical example, the Fokker–Planck equation can be formulated within the EnVarA framework through an energy–dissipation law. It is well known that the Fokker-Planck equation \eqref{eq:fpe} satisfies an energy-dissipation law as follows:
\begin{equation}\label{eqn:EnergyLaw_generalizedDiff}
    \frac{d E[\rho]}{dt}=-\int_\Omega\rho|\bm v|^2d\bm x,
\end{equation}
and can be considered as the continuity equation for the probability density $\rho$
\begin{equation} \label{eqn:diffusion}
\rho_t + \nabla \cdot (\rho {\bm v}) = 0.
\end{equation}
Here, ${\bm v}$ is the average velocity of all stochastic trajectories  given by
\begin{equation}\label{eqn:energy}
  {\bm v} = -\nabla (K_BT\ln\rho + \psi ).
\end{equation}

Using the general framework of EnVarA \cite{EnVarA1} mentioned above, we derive the Fokker-Planck equation \eqref{eq:fpe} from the energy-dissipation law (\ref{eqn:EnergyLaw_generalizedDiff}).   Note that the kinetic energy $\mathcal{K}$, the free energy $\mathcal{F}[f]$ and the dissipation functional $\mathcal{D}$ are given by
\begin{equation}
      \mathcal{K}=0, \qquad \mathcal{F} = \int_\Omega \left[K_BT\rho\ln\rho+\psi \rho\right]d\bm x,\qquad
      \mathcal{D}=\frac{1}{2}\int_\Omega\rho|\bm v|^2d\bm x\nonumber.
\end{equation} 
For the LAP, we introduce the flow map $\x(\X, t)$, defined as
\begin{equation} \label{eqn:ODE_flow_map}
 \begin{cases}
 & \frac{\dd}{\dd t} \x(\X, t) = {\bm v}(\x(\X, t), t), \\
 & \x(\X, 0) = \X,\nonumber
 \end{cases}
\end{equation}
for a given velocity field ${\bm v}$.
Here, $\X$ is the Lagrangian coordinate and $\x$ is the Eulerian coordinate. For fixed $\X$, $\x(\X, t)$ can be viewed as the trajectory of the particle that starts from the location $\X$. Due to mass conservation, $\rho(\x, t)$ can be considered as a function of the flow map $x(\X, t)$
\begin{equation}
  \rho(\x, t) =  \rho_0 (\X) / \det ( \nabla_{X} \x(\X, t) )\nonumber,
\end{equation}
where $\rho_0(\X)$ is the initial density. Taking the variation of the action functional with respect to the flow map $\x(\X, t)$, we obtain
the final force balance equation
\begin{equation}\label{eq:FPE-forcebalance}
\rho(\x,t){\bm v}(\x, t) = -\rho(\x,t)\left[K_BT\nabla\ln\rho(\x,t)+\nabla\psi(\x)\right],
\end{equation}
and thus 
\begin{equation}
{\bm v}(\x, t) = -\left[K_BT\nabla\ln\rho(\x,t)+\nabla\psi(\x)\right]\nonumber,
\end{equation}
which is the velocity derived from the energy-dissipation law (\ref{eqn:EnergyLaw_generalizedDiff}).
Together with the continuity equation~\eqref{eqn:diffusion}, we obtain the Fokker--Planck equation \eqref{eq:fpe}. 
Deriving the governing equation from an energy-dissipation law has the advantage that the resulting system is automatically thermodynamically consistent, i.e. it satisfies the fluctuation-dissipation theorem. For more details, we refer to the work \cite{EnVarA1, hu2024energetic} and the references therein.

This energetic variational structure, together with the associated variational formulations of the action and dissipation functionals, provides the foundation for the loss functions developed in this work, which are constructed directly from the underlying energy–dissipation law.
\begin{remark}
    As seen from the derivation of the Fokker–Planck equation via the EnVarA, this framework consists of two main components: the specification of the kinematics \eqref{eqn:diffusion} and the derivation of the force balance equation. The latter is obtained through two distinct variational principles, namely the LAP and the MDP. In other words, the underlying generalized diffusion \eqref{eq:sde} is fully characterized by two fundamental ingredients: the prescribed kinematics \eqref{eqn:diffusion}, which encode conservation laws such as mass conservation, and the force balance equation \eqref{eq:FPE-forcebalance}. In Section \ref{sec:method}, we will show how this structural insight can be leveraged to construct an effective loss function for learning unknown potential functions in the generalized diffusion model \eqref{eq:sde}.
\end{remark}

\subsection{A Brief Review of Related Works}
A wide range of methods can be employed to learn the potential function of the gradient system \eqref{eq:sde}, including probabilistic and statistical inference approaches, PDE-based methods, and energy-based methods, among others, as discussed in Section \ref{sec:intro}. These different approaches give rise to distinct loss functions and impose different requirements on the observational data. To set the stage for presenting our learning framework and to facilitate the discussion of its connection with PDE-based methods in the next section, we briefly review a basic form of the PDE-based approach from the literature and summarize our previous work.

\paragraph{PDE-based Learning Framework} To learn the potential function $\psi$ in the Fokker–Planck equation \eqref{eq:fpe} from observations of the solution $\rho$, one may solve the following PDE-based optimization problem to find the estimation of $\psi$:
\begin{equation}\label{eq:PDE_loss}
    \hat{\psi}=\argmin_{\psi}\left\|\rho_t-\nabla \cdot \left( \frac{1}{\gamma}  \rho \nabla (k_B T \ln \rho + \psi ) \right)\right\|^2,
\end{equation}
where $\|\cdot\|$ denotes an appropriate norm. Various advanced strategies have been proposed to enhance the effectiveness of PDE-based learning frameworks; see, for example, \cite{PINN_review}. For the sake of clarity and comparison with our approach in the next section, we restrict our discussion to this simplest yet representative form of the PDE-based loss function \eqref{eq:PDE_loss}.

\paragraph{A Differential-Form Loss of EnVarA Learning Framework} In our previous work \cite{lu2025EnVarA}, we proposed to learn the potential function via minimizing the following loss function 
\begin{equation}\label{eq:diffForm}
    \hat{\psi}=\argmin_{\psi}\left|\frac{dE}{dt}+\int_{\Omega}\rho|K_BT\nabla\ln\rho+\nabla\psi|^2d\bm x\right|^2.
\end{equation}
Recalling the energy–dissipation law \eqref{ED_1} associated with the gradient system \eqref{eq:sde}, it is not difficult to see that the idea underlying our work shares the same spirit as physics-informed neural networks (PINNs) \cite{PINN}, namely, learning the potential function by minimizing the residual of a governing physical law. In PDE-based approaches, the potential is typically learned by minimizing a pointwise loss function that enforces the underlying PDE at individual space–time locations.

In contrast, the energy-based method \eqref{eq:diffForm} considered here relies on the energy–dissipation law, which is a spatial-integral form. This fundamental difference leads to distinct requirements on the observational data. PINN-based methods are generally more suitable for relatively high-quality data and place fewer demands on repeated observations of the system's temporal evolution. On the other hand, the energy-based approach is more appropriate for noisier or lower-quality data but requires repeated observations over time to enhance identifiability, since the spatial-integral form yields a scalar relation and therefore benefits from greater data diversity. We refer to the loss function in \eqref{eq:diffForm} as the differential-form loss function, since it involves time derivatives. This terminology is adopted to distinguish it from the time-integral form introduced in the forthcoming section.

\section{Methodology}\label{sec:method}
In our previous work \cite{lu2025EnVarA},  we proposed minimizing the residual of the energy–dissipation law $\frac{dE}{dt}=-2\mathcal{D}$ at a single time instant to learn the unknown potential function from observational data. However, this loss formulation \eqref{eq:diffForm} lacks robustness with respect to the choice of the observation time. In particular, it works well when the data are collected sufficiently close to steady state(s) in order to enhance the identifiability of the potential function. 

In this work, we propose two strategies that not only alleviate the reliance on observation time but also incorporate richer structural information into the learning process.
\subsection{An Integral Form of EnVarA Learning Framework}\label{sec:integralForm}
A natural extension is to integrate the differential form of the energy–dissipation law with respect to time, yielding $E(T_e)-E(T_b)=-2\int_{T_b}^{T_e}\mathcal{D}(t)dt$ in a given time interval $[T_b,T_e]$. By minimizing the residual of this integral form of the energy–dissipation law
\begin{equation}\label{eq:integralForm}
    \hat{\psi}=\argmin_{\psi}\left|E[T_e]-E[T_b]+\int_{T_b}^{T_e}\int_{\Omega}\rho|K_BT\nabla\ln\rho+\nabla\psi|^2d\bm xdt\right|^2\nonumber,
\end{equation}
we are able to learn the potential function from data that are not necessarily close to steady state(s). We will demonstrate later that this extension significantly enhances robustness with respect to observation times.

\subsection{The De Giorgi-Type Energy-Dissipation Functional for Generalized Diffusion Systems}{Generalized Diffusion Systems}
In addition to the natural extension discussed in Subsection~\ref{sec:integralForm}, we introduce in this subsection a De Giorgi-type formulation to further improve the EnVarA-based learning framework.

\subsubsection{The Variational Description of Generalized Diffusion Systems}

{
Generalized diffusions are models that can be characterized
by the energy--dissipation law
\begin{align}\label{eq:H-1_velocity}
    \frac{dE}{dt}
    =-2\mathcal{D}
    =-\int_{\Omega}\rho|\bm v|^2\,d\bm x,
\end{align}
together with the kinematic relation
\begin{equation}\label{kinematics}
    \rho_t+\nabla\cdot(\rho\bm v)=0.
\end{equation}
Here, \(E=E[\rho]\) denotes the free energy, \(\bm v\) is the transport
velocity, and \(\mathcal{D}\) is the dissipation functional. We assume that
the dissipation is quadratic in \(\bm v\), corresponding to the
linear-response regime \cite{kubo1966fluctuation}.

Let
\[
    \mu=\frac{\delta E}{\delta\rho}
\]
denote the chemical potential. Applying the least action principle and the
maximum dissipation principle yields the force-balance relation
\begin{equation}\label{eq:H-1_forcebalance}
    -\rho\bm v=\rho\nabla\mu.
\end{equation}
Combining \eqref{eq:H-1_forcebalance} with the mass-conservation law
\eqref{kinematics}, we obtain
\begin{equation}
    \rho_t=\nabla\cdot(\rho\nabla\mu).
\end{equation}
Moreover, substituting the force-balance relation
\eqref{eq:H-1_forcebalance} into \eqref{eq:H-1_velocity} gives the equivalent
chemical-potential form of the energy--dissipation law:
\begin{align}\label{eq:H-1_chemicalPotential}
    \frac{dE}{dt}
    =-\int_{\Omega}\rho|\nabla\mu|^2\,d\bm x.
\end{align}
Indeed, for any \(\alpha\in[0,1]\), we can write the energy-dissipation law in a more general form
\begin{align}\label{eq:H-1_cheeger-diffForm}
    \frac{dE}{dt}&=-\int_{\Omega} \left[ (1-\alpha)\rho|\bm v|^2+\alpha\rho|\nabla\mu|^2 \right] d\bm x.
\end{align} 
The integral form of the energy-dissipation law \eqref{eq:H-1_cheeger-diffForm} is the following
\begin{align}\label{eq:H-1_cheeger}
    E(T_e)=E(T_b)-\int_{T_b}^{T_e}\int_{\Omega} \left[(1-\alpha)\rho|\bm v|^2+\alpha\rho|\nabla\mu|^2 \right] d\bm xdt.
\end{align}
}
\begin{remark}
     For clarification, we emphasize that the EnVarA is a fairly general framework for deriving the governing equations of dissipative systems by prescribing the free energy, the dissipation rate, and the kinematics. In Section \ref{sec:EnVarA}, taking the Fokker–Planck equation as an example, we derive the governing equation from the energy–dissipation law \eqref{eq:H-1_velocity}. In addition, the energy dissipation law \eqref{eq:H-1_cheeger} with $\alpha=\frac{1}{2}$ is commonly referred to as the \emph{De Giorgi-type energy-dissipation functional} \cite{deGiorgi1,deGiorgi2,ambrosio2005gradient}. It is worth noting that the resulting functional no longer admits a direct physical interpretation as a dissipation functional, as it is not uniquely determined by the underlying flow map. Nevertheless, although this reformulation is purely algebraic, it remains useful for learning the potential function.
\end{remark}
In particular, when $\alpha=\frac{1}{2}$, the energy--dissipation law \eqref{eq:H-1_cheeger-diffForm} (equivalently, the De Giorgi-type energy-dissipation functional \eqref{eq:H-1_cheeger}) is equivalent to the \emph{velocity-matching} condition
\begin{equation}\label{eq:velocity-matching}
    \bm v=-\nabla\mu \quad \text{a.e.\ on }\{\rho>0\},
\end{equation}
i.e., the force-balance relation \eqref{eq:H-1_forcebalance}. To see this, we directly compute
\begin{align} \label{eq:equiv1}
& \frac{dE}{dt} + \int_{\Omega} \frac{1}{2} \big( \rho |\bm v|^2 + \rho |\nabla \mu|^2 \big) \, d\bm x
= \int_{\Omega} \mu \rho_t \, d\bm x
+ \int_{\Omega} \frac{1}{2} \big( \rho |\bm v|^2 + \rho |\nabla \mu|^2 \big) \, d\bm x \nonumber\\
&= -\int_{\Omega} \mu \nabla \cdot (\rho \bm v)\, d\bm x
+ \int_{\Omega} \frac{1}{2} \big( \rho |\bm v|^2 + \rho |\nabla \mu|^2 \big) \, d\bm x \nonumber = \int_{\Omega} \rho \bm v \cdot \nabla \mu \, d\bm x
+ \int_{\Omega} \frac{1}{2} \big( \rho |\bm v|^2 + \rho |\nabla \mu|^2 \big) \, d\bm x \nonumber\\
&= \int_{\Omega} \frac{1}{2} \rho |\bm v + \nabla \mu|^2 \, d\bm x.
\end{align}
The equivalence is crucial for establishing our learning framework for identifying the unknown potential function $\psi$, and we will revisit it later in the discussion of the learning framework.

\subsubsection{\texorpdfstring{A Special Case of Generalized Diffusion System: the Fokker-Planck Equation}{}}
The Fokker-Planck equation~\eqref{eq:fpe} can be written as a continuity equation:
\begin{align} \label{eq:fpec}
    \frac{\partial\rho}{\partial t}=\nabla\cdot(\rho\nabla[K_BT\ln\rho+\psi])=-\nabla\cdot(\rho\bm v),\quad \text{where the velocity}\quad \bm v=-\nabla(K_BT\ln\rho+\psi).
\end{align}

The free energy of the Fokker-Planck equation is defined as
\begin{equation} \label{eq:EforFPE}
    E[\rho]=\int_{\Omega} (K_BT\rho\ln\rho+\rho\psi)\, d\bm x,
\end{equation}
where $\psi$ is the potential function and the chemical potential is $\mu=\frac{\delta E}{\delta\rho}=K_BT\ln\rho+\psi$.

As in \eqref{eq:H-1_cheeger-diffForm}, the energy-dissipation law of the Fokker-Planck equation can also be written as 
\begin{equation}\label{eq:H-1_cheeger_FP}
     \frac{dE}{dt}=-\int_{\Omega} \left[ (1-\alpha)\rho|\bm v|^2+\alpha\rho|\nabla\mu|^2\right] \, d\bm x\nonumber.
\end{equation}
It should be noted that here we specify the free energy $E$ as defined in \eqref{eq:EforFPE}, i.e. $E[\rho]=\int_{\Omega} (K_BT\rho\ln\rho+\rho\psi)\, d\bm x$ while it is not specified for the description of generalized diffusion systems in the previous subsection.

\subsubsection{Loss Functions}
In this subsection, we present three different loss functions for learning the unknown potential function $\psi$ in the generalized diffusion \eqref{eq:sde}, each derived from one of the three energy–dissipation laws associated with the Fokker–Planck equation \eqref{eq:fpec} described above. We then discuss the key distinctions among these loss functions.

Suppose we aim to learn the unknown potential function $\psi$ from a given dataset on $\rho$ or $\rho$ and $\bm v$. As for which dataset should be used, it depends on the choice of loss function, and we will emphasize this later. To this end, we construct a neural network to approximate the unknown potential function, denoted by $\psi^{NN}$. In order to find the 'best' parameters $\theta^*$ of the neural network, we can minimize the following loss function
\begin{equation}
    \theta^*=\argmin_{\theta} L_{\text{energy}}^{\alpha}\nonumber,
\end{equation}
where the loss function $L_{\text{energy}}^{\alpha}$ is defined by
\begin{align} \label{eq:lossfuncci}
    L_{\text{energy}}^{\alpha} = \left(E^{NN}(T_e)-E^{NN}(T_b)+\int_{T_b}^{T_e}\int_{\Omega} \left[ (1-\alpha)\rho|\bm v|^2+\alpha\rho|\nabla(K_BT\ln\rho+\psi^{NN})|^2\right]\, d\bm xdt\right)^2.
\end{align}
 The notation $E^{NN}$ means that the free energy is approximated by the neural network $\psi^{NN}$, that is $E^{NN}=\int_{\Omega} (\rho\ln\rho+\rho\psi^{NN})\, d\bm x$.  In this work, we mainly focus on three representative cases, namely $\alpha=0,0.5,1$. If $\alpha=0$, the loss function becomes
 \begin{equation}
    L_{\text{energy}}^{\alpha=0} = \left|E^{NN}(T_e)-E^{NN}(T_b)+\int_{T_b}^{T_e}\int_{\Omega}\rho|\bm v|^2d\bm xdt\right|^2\nonumber.
\end{equation}
For this case, the dissipation rate $\int_{\Omega}\rho|\bm v|^2dx$ can be determined by observational data on $\rho$ and $\bm v$.

If $\alpha=1$, the loss function becomes
\begin{align}
    L_{\text{energy}}^{\alpha=1} = \left|E^{NN}(T_e)-E^{NN}(T_b)+\int_{T_b}^{T_e}\int_{\Omega}\rho|\nabla(K_BT\ln\rho+\psi^{NN})|^2d\bm xdt\right|^2\nonumber,
\end{align}
which implies that we do not need the observation on the velocity $\bm v$ and this corresponds to the integral-form version of the loss function \eqref{eq:diffForm} that was proposed in our previous work \cite{lu2025EnVarA}. 

Similarly, when $\alpha=0.5$, the term $\int_{\Omega}\rho|\bm v|^2d\bm x$ can be determined by observational data while the term $\int_{\Omega}\rho|\nabla(K_BT\ln\rho+\psi^{NN})|^2d\bm x$ involves the neural network $\psi^{NN}$. Thus, this case still needs to use the observation on both $\rho$ and $\bm v$.

\begin{align} \label{eq:lossfuncci_alpha=0.5}
    L_{\text{energy}}^{1/2} = \left(E^{NN}(T_e)-E^{NN}(T_b)+\int_{T_b}^{T_e}\int_{\Omega} \left[\frac{1}{2}\rho|\bm v|^2+\frac{1}{2}\rho|\nabla(K_BT\ln\rho+\psi^{NN})|^2\right]\, d\bm xdt\right)^2\nonumber.
\end{align}


\subsection{From SDE Snapshots to Discrete Energy–Dissipation Loss}
In practice, the density $\rho$ may not be directly observable. Instead, we assume access to the particle data corresponding to the particles satisfying the  gradient system~\eqref{eq:sde}. In particular, the observational data are the particle positions $\left\{\bm x_j^q(s_i)\right\}_{i,j,q=1}^{M,N,Q}$ or the positions and velocities $\left\{\bm x_j^q(s_i), \bm v_j^q(s_i)\right\}_{i,j,q=1}^{M,N,Q}$, where the time instances $s_i\in[T_b,T_e], i=1,2,\ldots,M$, are uniformly spaced with time step size $\Delta s$. Here, the index $j$ labels individual particles, while the index $q$ corresponds to different initial distributions, with $N$ particles sampled from each initial distribution. In other words, we repeatedly simulate or observe independent ensembles of the system~\eqref{eq:sde}, where each ensemble is initialized from a distinct distribution and consists of $N$ sampled particles.
To avoid explicit computation of spatial integrals in the energy–dissipation law, we employ a particle-based discretization.

The density $\rho(\bm x,t)$ is approximated by  $\rho_N(\bm x,t)=\frac{1}{N}\sum\limits_{j=1}^N\delta(\bm x-\bm x_j(t))$. Therefore, the energy $E$ is discretized as 
\begin{equation} \label{eq:energyd}
    E[\rho](t)=\int_{\Omega} (K_BT\rho\ln\rho+\rho\psi) \,d\bm x
    \approx \frac{1}{N}\sum\limits_{j=1}^N[K_BT\ln(K_h*\rho_N)(\bm x_j)+\psi(\bm x_j)],
\end{equation}
where $(K_h*\rho_N)(\bm x)=\int_{\Omega} K_h(\bm x, \bm y)\rho_N(\bm y,t)d\bm y=\frac{1}{N}\sum\limits_{j=1}^NK_h(\bm x,\bm x_j(t))$ and $h$ denotes the bandwidth of the kernel $K_h$. For $K_h$, we use the Gaussian kernel defined as $K_h(x,y)=C_{h}^{-1}\exp\{-\frac{\|x-y\|^2}{2h^2}\}$ with the normalization constant $C_h=(2\pi)^{\frac{n}{2}}h^{n}$.

Similarly, the dissipation rate can be discretized as 
\begin{align} \label{eq:dissrated}
    \mathcal{D}=\frac{1}{2}\int_{\Omega}\rho|K_BT\nabla\ln\rho+\nabla\psi|^2 \,d\bm x
    \approx\frac{1}{2N}\sum\limits_{j=1}^N|K_BT\nabla\ln[(K_h*\rho_N)(\bm x_j)]+\nabla\psi(\bm x_j)|^2.
\end{align}

From the two equations~\eqref{eq:energyd} and \eqref{eq:dissrated}, we discretize the energy-dissipation law~\eqref{eq:H-1_cheeger} as
\begin{align}\label{eq:disenlawd}
 &\frac{1}{N}\sum\limits_{j=1}^N\{K_BT\ln[(K_h*\rho_N)(\bm x_j^q(T_e))]+\psi^{NN}(\bm x_j^q(T_e))\} -  \frac{1}{N}\sum\limits_{j=1}^N\{K_BT\ln[(K_h*\rho_N)(\bm x_j^q(T_b))]+\psi^{NN}(\bm x_j^q(T_b))\} \nonumber\\
 & \approx -2\left[\frac{1-\alpha}{2N}\sum\limits_{i,j=1}^{M,N}|\bm v_j^q(s_i)|^2+\frac{\alpha}{2N}\sum\limits_{i,j=1}^{M,N}|K_BT\nabla\ln[(K_h*\rho_N)(\bm x_j^q(s_i))]+\nabla\psi^{NN}(\bm x_j^q(s_i))|^2\right] \Delta s.
\end{align}

We propose to minimize the residual of the discretized energy-dissipation law~\eqref{eq:disenlawd} to learn the unknown potential function $\psi$. The discretization of the loss function \eqref{eq:lossfuncci} can be written as follows
\begin{align}\label{eq:discretizedEnergyloss}
L_d= {} & \sum_{q=1}^Q\Bigg|
\frac{1}{N}\sum_{j=1}^N \Big[ K_BT\ln[(K_h*\rho_N)(\bm x_j^q(T_e))]
+ \psi^{NN}(\bm x_j^q(T_e)) \Big] \nonumber \\
& \quad
- \frac{1}{N}\sum_{j=1}^N \Big[ K_BT\ln[(K_h*\rho_N)(\bm x_j^q(T_b))]
+ \psi^{NN}(\bm x_j^q(T_b)) \Big]  \nonumber \\
& \quad
+ 2 \left[\frac{1-\alpha}{2N}
\sum_{i,j=1}^{M,N} \big| \bm v_j^q(s_i) \big|^2
+ \frac{\alpha}{2N}
\sum_{i,j=1}^{M,N}
\big| K_BT\nabla\ln[(K_h*\rho_N)(\bm x_j^q(s_i))]
+ \nabla\psi^{NN}(\bm x_j^q(s_i)) \big|^2
\,\right] \Delta s
\Bigg|^2 .
\end{align}

\begin{remark}\label{rmk:direct}
    The loss function is constructed by minimizing the residual of the discretized energy-dissipation law. Recall from the De Giorgi-type energy-dissipation functional~\eqref{eq:equiv1} that
\begin{align}
    \frac{dE}{dt} + \int_{\Omega}\frac{1}{2}\big(\rho|\bm v|^2+\rho|\nabla\mu|^2\big)\,d\bm x
    =\int_{\Omega}\frac{1}{2}\rho|\bm v+\nabla\mu|^2\,d\bm x \nonumber,
\end{align}
so that the continuous energy-dissipation law for $\alpha=\frac{1}{2}$, namely $\frac{dE}{dt}+\int_{\Omega}\frac{1}{2}(\rho|\bm v|^2+\rho|\nabla\mu|^2)\,d\bm x=0$, holds if and only if the right-hand side above vanishes. The residual of the continuous version of the energy-dissipation law is thus
\begin{align}\label{eq:loworder_pde_loss}
    L_{\text{PDE}}=\int_{\Omega}\frac{1}{2}\rho(\bm v+\nabla\mu)^2 \,d\bm x
    =\int_{\Omega}\frac{1}{2}\rho(\bm v+K_BT\nabla\ln\rho+\nabla\psi)^2\, d\bm x.
\end{align}
This corresponds to a low-order PDE-based loss function, as it avoids minimizing the residual of the full continuity equation $\rho_t=-\nabla\cdot(\rho\bm v)$ by leveraging direct observations of the velocity field $\bm v$. One may then ask why the PDE-based loss function \eqref{eq:loworder_pde_loss} is not used instead of the loss function \eqref{eq:discretizedEnergyloss}.
In fact, the loss function \eqref{eq:discretizedEnergyloss} is more robust than the PDE-based loss \eqref{eq:loworder_pde_loss}, as it preserves richer dynamical information. More specifically, the loss in \eqref{eq:discretizedEnergyloss} guides the learning process toward a potential function that maintains the energy–dissipation structure of the underlying system as much as possible.
\end{remark}

\begin{remark}
The kernel-smoothed density 
$$
(K_h * \rho_N)(\bm x) =  \frac{1}{N}\sum_{j=1}^N K_h(\bm x,\bm x_j(t))
$$
is introduced to regularize the empirical particle density.
This smoothing step avoids the numerical difficulties associated with evaluating spatial derivatives and the logarithmic function of the empirical measure.
We note that the choice of the kernel bandwidth $h$ may become challenging in high-dimensional settings and requires further investigation.
However, addressing this issue is beyond the scope of the present work, as our numerical experiments are restricted to problems in one to three dimensions.
\end{remark}

\section{Numerical Results}
In this section, we first compare the learning effectiveness of three loss functions~\eqref{eq:lossfuncci} or \eqref{eq:discretizedEnergyloss} corresponding to $\alpha=0, 0.5$ and $1$ using a simple one-dimensional (1D) example. We then investigate the robustness of the loss function based on the De Giorgi-type energy-dissipation functional ($\alpha=0.5$) under different experimental settings through two- and three-dimensional examples.

In what follows, we set the parameters in the stochastic differential equation \eqref{eq:sde} as $\gamma=1$, and $K_BT=0.125$. For all numerical experiments, the unknown potential $\psi$ is approximated using a fully connected neural network with a single hidden layer consisting of 64 neurons. The {\bf Tanh()} function is used as the activation function, and the network is trained using the \emph{Adam} optimizer with a learning rate of $1\times 10^{-3}$. Training is performed for 5,000 epochs in all cases.

\subsection{A 1D Example}\label{sec:1d}
\begin{example}
    In the first example, we aim to test the effectiveness of the three loss functions~\eqref{eq:discretizedEnergyloss} corresponding to $\alpha=0,0.5$ and $1$. This one-dimensional example serves as a minimal testbed that allows us to clearly illustrate the role of the parameter $\alpha$ without the additional complexity introduced by higher-dimensional dynamics. 
    
    The target potential is chosen as $\psi(x)=\frac{1}{4}x^4-\frac{1}{2}x^2$, which is a prototypical double-well potential. This choice gives rise to two stable metastable states separated by an energy barrier, making it a canonical model for bistable systems and noise-induced transitions. Such potentials are widely used to assess the ability of learning methods to capture nontrivial energetic landscapes and barrier-crossing dynamics in stochastic gradient systems. We construct $Q=5$ Gaussian initial distributions with means uniformly sampled from $[-2,2]$ and a common standard deviation of $0.5$. For each initial distribution, we simulate the associated SDE~\eqref{eq:sde} using $N=2,000$ particles to generate sample trajectories. The system states are recorded at $M=6$ equally spaced time points over the interval $[0.3,0.8]$ with a fixed time step of 0.1. The velocity field $v$ is computed by $v=-\nabla\mu\approx-[K_BT\nabla\ln(K_h*\rho_N)+\nabla\psi]$, where the relation comes from the force balance equation $\rho v=-\rho\nabla\mu$ and the bandwidth $h=0.05$. These samples on $\rho$ and $v$ collectively form the training dataset. In practice, the velocity $v$ may need to be obtained directly from observations, rather than inferred through the force balance relation. The feasibility and methodology for measuring or estimating this quantity depend strongly on the specific application domain and the available experimental or observational techniques. Since such issues are highly problem-dependent and lie beyond the scope of the present work, we do not pursue them further here. 

\begin{figure}[htbp]
    \centering
        \includegraphics[width=0.4\textwidth]{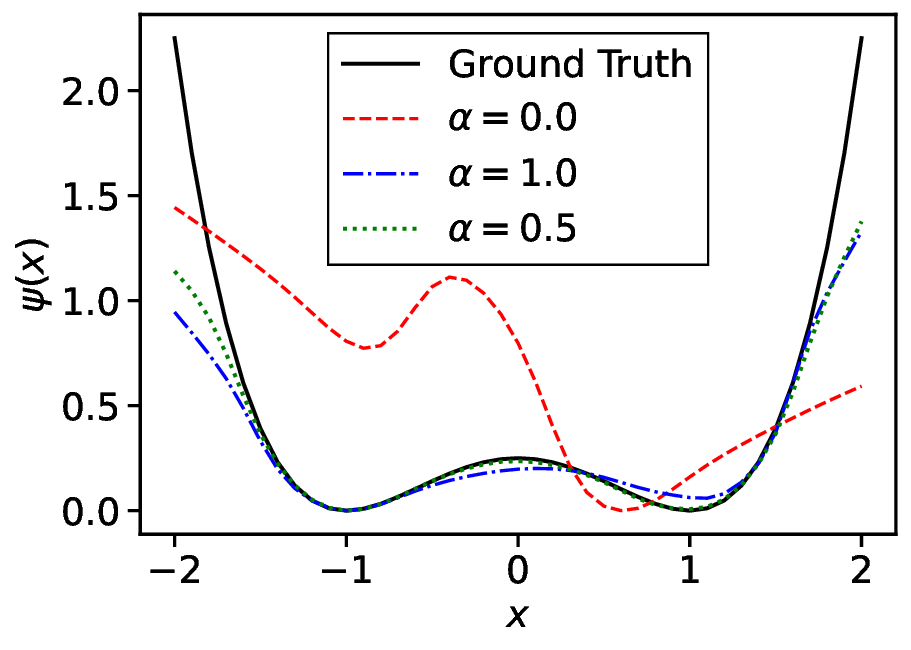}
    \caption{\it Learning the potential $\psi(x)=\frac{1}{4}x^4-\frac{1}{2}x^2$ using the loss function \eqref{eq:discretizedEnergyloss} for different values of the weighting parameter $\alpha$.}
    \label{fig:1dEx_double-well}
\end{figure}

    {As shown in Fig.~\ref{fig:1dEx_double-well}, the learned potentials with $\alpha=0.5$ and $\alpha=1$ recover the major features of the target potential $\psi$, particularly the locations and structures of the two potential wells, whereas the case with $\alpha=0$ fails to identify the target potential from the available observational data.} Indeed, for the loss function \eqref{eq:discretizedEnergyloss} with $\alpha=0$, the dissipation rate is obtained directly from the observational data and does not change in the training process. Moreover, when $\alpha=0$, the unknown potential only appears in the calculation of the free energy \eqref{eq:energyd} and the overall loss function~\eqref{eq:discretizedEnergyloss} does not provide the gradient information of the unknown potential compared to the cases $\alpha=0.5$  and $\alpha=1$. Thus, the loss function~\eqref{eq:discretizedEnergyloss}  with $\alpha=0$ does not provide an effective way to learn the potential function due to the integral-form loss and lack of the gradient information of the unknown potential function in the loss function. We note that, although some discrepancies remain near the boundary regions of the computational domain, these deviations can be further reduced by increasing the spatial coverage of the observational data. Nevertheless, accurately recovering such boundary details is not our primary focus. Under the constraint of limited observations, robust identification of the dominant potential landscape is more critical for characterizing the long-time behavior of the underlying dynamical system.
      
    Although both $\alpha=0.5$ and $\alpha=1$ recover the major features of the target potential (i.e., the double-well structure) in this 1D setting, the training data of the case $\alpha=1$ relies only on the observational data of $\rho$ while the case $\alpha=0.5$ relies on the data on $(\rho,v)$. From the formulation of the loss function~\eqref{eq:discretizedEnergyloss}, one might expect the choice of $\alpha=1$ to be advantageous, as it directly exploits the full gradient information of the potential, while the corresponding contribution is only weighted by $0.5$ for $\alpha=0.5$. As shown in Fig.~\ref{fig:1dEx_double-well}, the two choices lead to comparable learning results in the 1D setting, both recovering the dominant features of the target potential. However, this advantage of $\alpha=1$ does not persist when the available data become limited. In particular, when the number of initial distributions $Q$ is reduced, the case with $\alpha=0.5$ consistently outperforms the one with $\alpha=1$. More pronounced differences between the two choices of $\alpha$ will be observed in higher-dimensional and noisy scenarios. This simple example highlights the necessity of incorporating dissipation-based gradient information and motivates the use of $\alpha=0.5$ in the subsequent higher-dimensional experiments.
\end{example}

\subsection{2D Examples}
\begin{example}
    Following the one-dimensional example, we next consider a two-dimensional generalized diffusion system with a quadruple-well potential given by
    \begin{equation}
    \psi(x,y)=\frac{1}{4}x^4-\frac{1}{2}x^2+\frac{1}{4}y^4-\frac{1}{2}y^2\nonumber.
    \end{equation}
    This potential is separable in the two spatial directions and gives rise to four symmetric metastable wells located at $(\pm 1, \pm 1)$. Compared with the one-dimensional double-well case, the two-dimensional setting introduces multiple transition pathways and a more intricate probability flow structure, providing a more stringent test for the proposed learning framework in higher dimensions. Although the potential is separable, the resulting diffusion dynamics exhibit nontrivial transport behavior in the joint $(x,y)$-space, making this example a meaningful intermediate benchmark between one-dimensional and fully coupled high-dimensional systems.
    
    Similar to the 1D example, we construct $Q$ Gaussian initial distributions with means uniformly sampled from the square $[-2,2]\times[-2,2]$ and a common standard deviation of $0.5$. For each initial distribution, we simulate the associated SDE~\eqref{eq:sde} using $N=10,000$ particles to generate sample trajectories. The system states are recorded at $M=6$ equally spaced time points over the interval $[T_b,T_e]$ with a fixed time step of $0.1$. The velocity data $\bm v$ are obtained by using the force balance equation again and the bandwidth $h$ of the kernel $K_h$ is still chosen as $0.05$.
    
    \paragraph{\bf{Comparison between the Two Choices $\alpha=0.5$ and $\alpha=1$ in the Loss Function~\eqref{eq:discretizedEnergyloss} }} In our previous work \cite{lu2025EnVarA}, the learning framework is based on the differential form of the energy dissipation law\eqref{eq:diffForm} and we observed that improved performance is achieved when training data are collected closer to equilibrium. Motivated by this observation, we compare the training performance by choosing three different time intervals, namely $[T_b, T_e]=[0.1, 0.6]$, $[0.3, 0.8]$ or $[0.5, 1.0]$, for both choices of the parameter $\alpha$ of the loss function~\eqref{eq:discretizedEnergyloss}: $\alpha = 0.5$ and $\alpha = 1$. For each time interval, neural networks are trained using the same simulated data under the parameter settings. In this example, the number of initial distributions is chosen as $Q=5$. Fig.~\ref{fig:2dEx_CP_5ICs} and Fig.~\ref{fig:2dEx_cheeger_5ICs} show the density plots of the learned potentials using the data from the time intervals $[T_b,T_e]=[0.1,0.6],[0.3,0.8]$ and $[0.5,1]$ for both $\alpha=1$ and $\alpha=0.5$, compared with the ground truth. Although the training data for both cases vary across different time intervals, the loss defined in \eqref{eq:discretizedEnergyloss} with $\alpha = 0.5$ consistently produces better and more robust performance than the case with $\alpha = 1$ for the same training time period.

    \begin{figure}[htbp]
    \centering
    \includegraphics[width=0.9\textwidth]{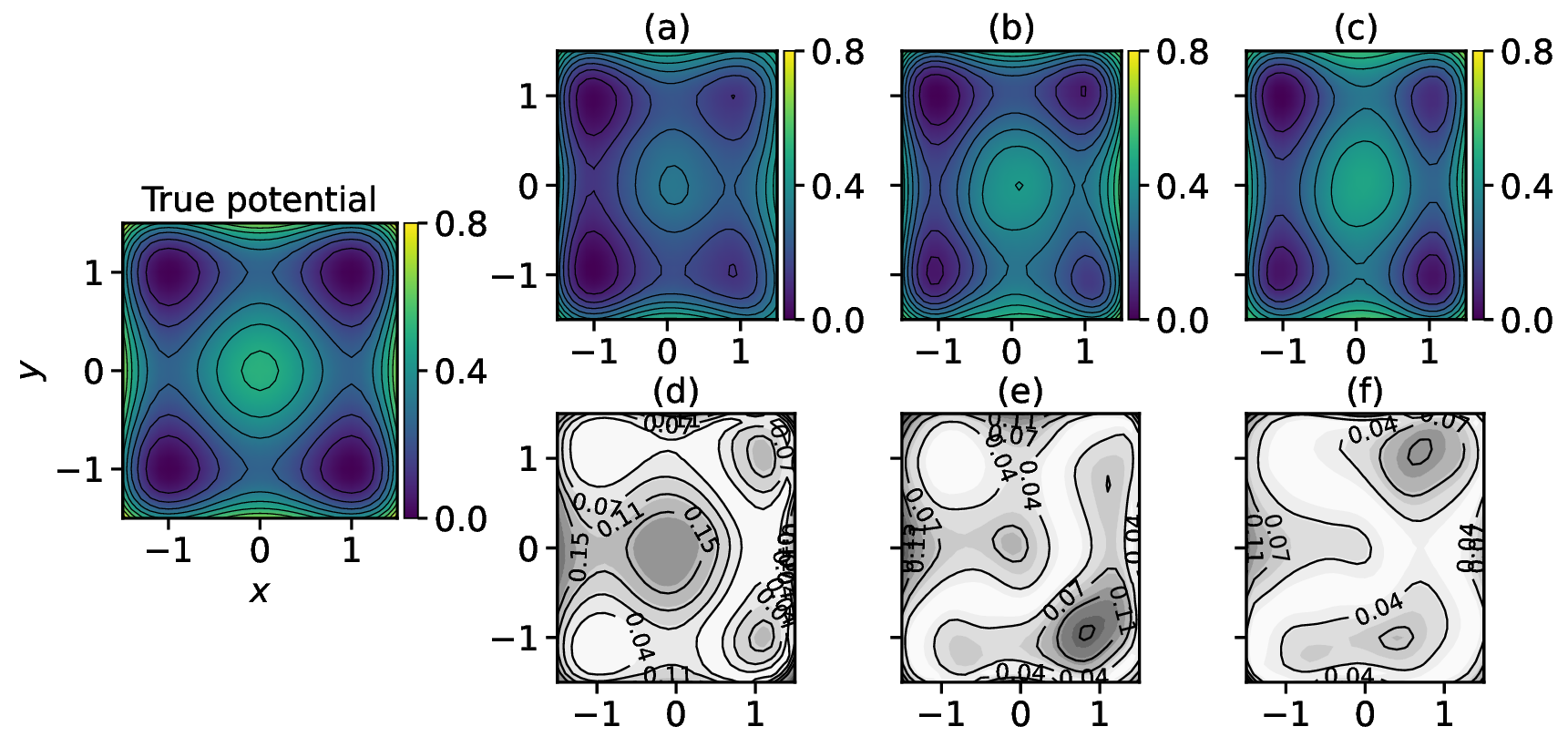}
    \caption{\it Learning the quadruple-well potential $\psi(x,y)=\frac{1}{4}x^4-\frac{1}{2}x^2+\frac{1}{4}y^4-\frac{1}{2}y^2$ from training data collected over different time intervals $[T_b,T_e]$, with $\alpha=1$ in the loss function~\eqref{eq:discretizedEnergyloss}. The leftmost panel shows the ground-truth potential. The first row presents the learned potentials using data collected over the time intervals $[0.1,0.6]$, $[0.3,0.8]$, and $[0.5,1.0]$, corresponding to panels (a)--(c), respectively. The second row displays the corresponding pointwise absolute errors between the learned potentials and the ground-truth potential, shown in panels (d)--(f).}
    \label{fig:2dEx_CP_5ICs}
    \end{figure}

    \begin{figure}[htbp]
    \centering
    \includegraphics[width=0.9\textwidth]{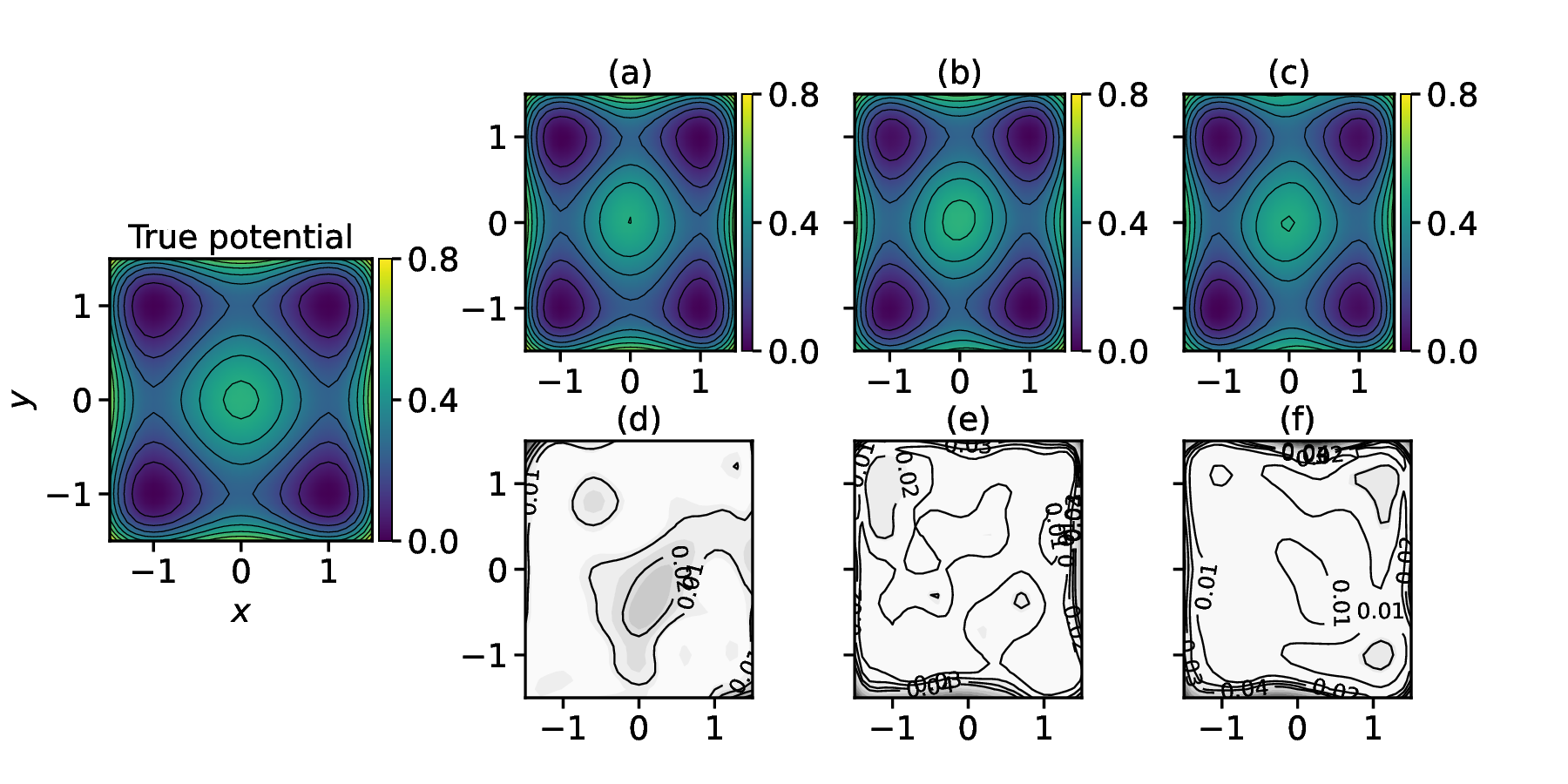}
    \caption{ \it Learning the quadruple-well potential $\psi(x,y)=\frac{1}{4}x^4-\frac{1}{2}x^2+\frac{1}{4}y^4-\frac{1}{2}y^2$ from training data collected over different time intervals $[T_b,T_e]$, with $\alpha=0.5$ in the loss function~\eqref{eq:discretizedEnergyloss}. The leftmost panel shows the ground-truth potential. The first row presents the learned potentials using data collected over the time intervals $[0.1,0.6]$, $[0.3,0.8]$, and $[0.5,1.0]$, corresponding to panels (a)--(c), respectively. The second row displays the corresponding pointwise absolute errors between the learned potentials and the ground-truth potential, shown in panels (d)--(f).}
    \label{fig:2dEx_cheeger_5ICs}
    \end{figure}

\paragraph{\bf Impact of the Number of Initial Distributions $Q$}
Fig.~\ref{fig:2dEx_CP_diffICs} shows the impact of the initial distributions $Q$ on the learning outcomes. When using the loss function \eqref{eq:discretizedEnergyloss} with $\alpha = 1$, we observe that the learning results are relatively inaccurate when only a small number of initial distributions is employed. To investigate whether this limitation can be mitigated, we examine the effect of increasing the number $Q$ of initial distributions used for training. Specifically, we consider $Q = 5$, $10$, and $20$ while keeping all other experimental settings fixed. In all cases, the training data are collected over the same time interval $[0.3,0.8]$. As illustrated in Fig.~\ref{fig:2dEx_CP_diffICs}, increasing $Q$ leads to a clear and consistent improvement in the learned potential. This observation indicates that, for $\alpha = 1$, the learning performance is sensitive to the diversity of the initial distributions, and a relatively large value of $Q$ is required to achieve satisfactory accuracy. In contrast, as shown in Fig.~\ref{fig:2dEx_cheeger_5ICs}, comparable accuracy can be achieved with fewer initial distributions (in this case, $Q=5$) when $\alpha = 0.5$, highlighting the improved robustness of this choice. Moreover, the improvement in Fig.~\ref{fig:2dEx_CP_diffICs} can also be understood from the perspective of spatial coverage of the training data. As the number of initial distributions $Q$ increases, the corresponding dissipative trajectories provide a more uniform exploration of the state space, allowing the learning framework to better capture the global structure of the underlying potential landscape. This effect is particularly pronounced for the present example, where the potential contains four symmetric and well-separated wells, and accurate recovery requires sufficient information from different regions of the energy landscape. These observations further indicate that, beyond the choice of the loss functional, structure-aware sampling strategies that enhance the coverage of dynamically relevant regions may provide an effective route toward more efficient learning of complex potential landscapes.

\begin{figure}[htbp]
\centering
\includegraphics[width=0.9\textwidth]{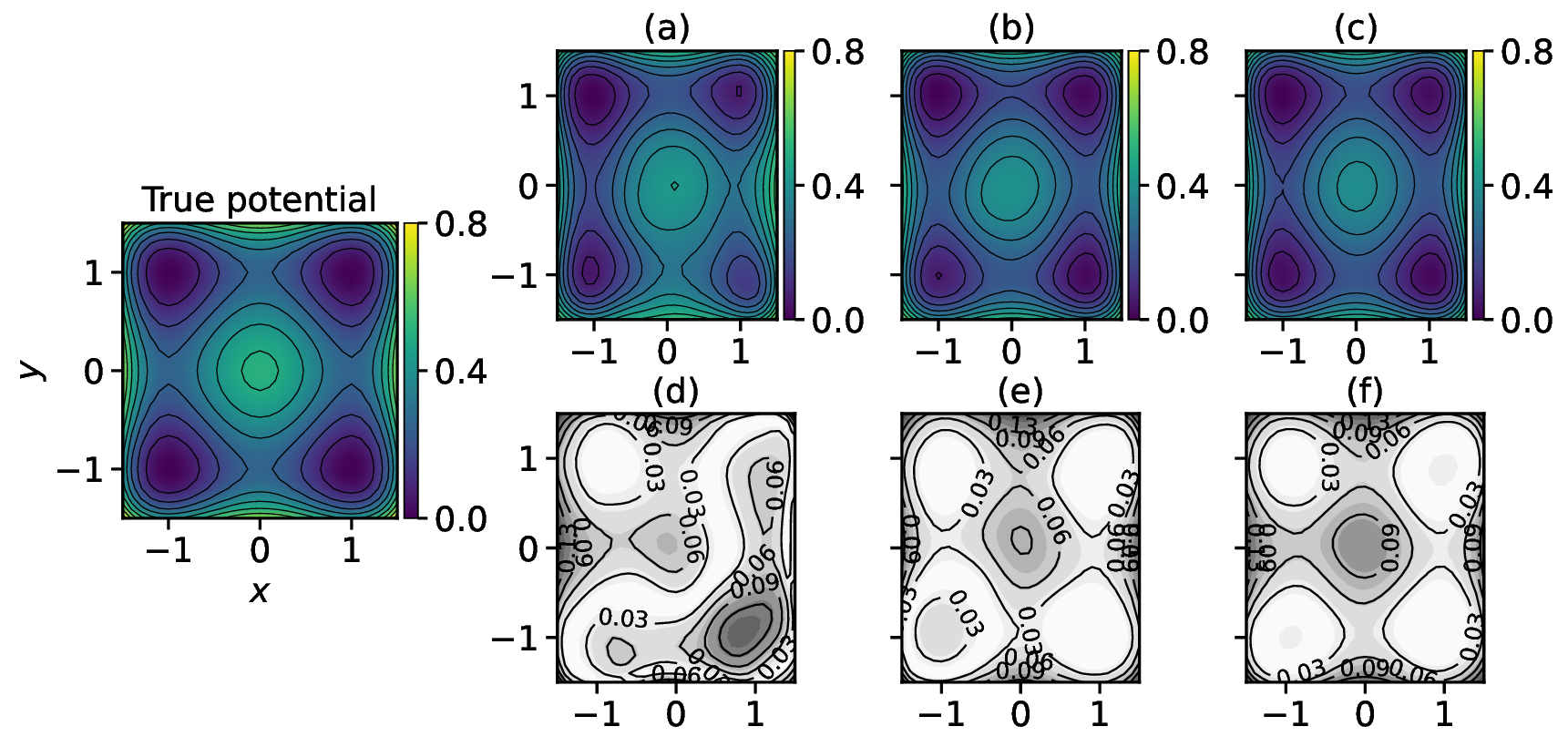}
\caption{\it Learning results with $\alpha=1$ using different numbers of initial distributions $Q$. The leftmost panel shows the ground-truth potential. Panels (a)--(c) present the learned potentials obtained with $Q=5$, $10$, and $20$ initial distributions, respectively, while panels (d)--(f) show the corresponding pointwise absolute errors between the learned potentials and the ground-truth potential.}
\label{fig:2dEx_CP_diffICs}
\end{figure}

\begin{figure}[htbp]
\centering
\includegraphics[width=0.9\textwidth]{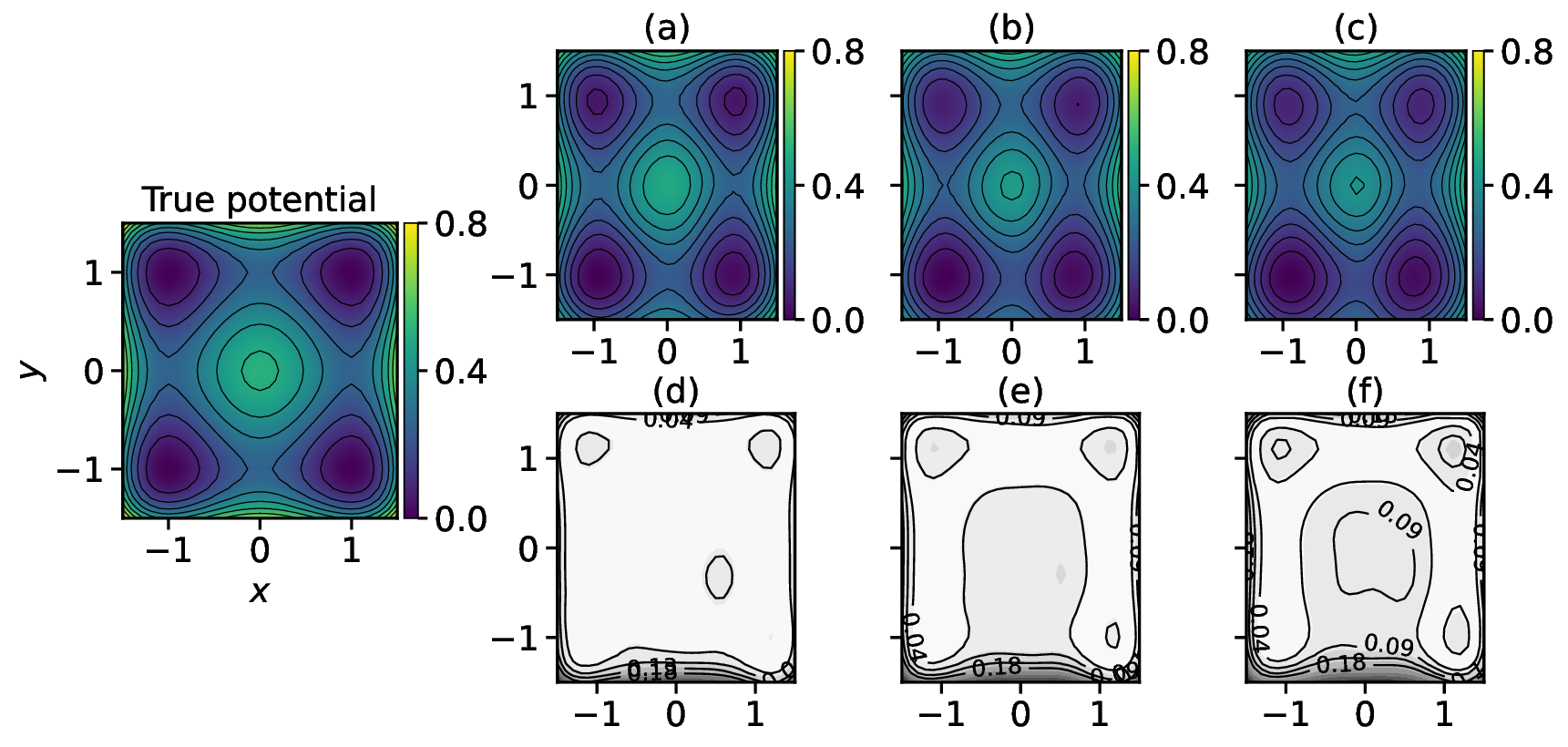}
\caption{\it Learning results with $\alpha=0.5$ under different noise levels $\sigma$. The leftmost panel shows the ground-truth potential. Panels (a)--(c) present the learned potentials obtained with noise level $\sigma=0.1$, $0.2$, and $0.3$, respectively, while panels (d)--(f) show the corresponding pointwise absolute errors between the learned potentials and the ground-truth potential.}
\label{fig:2dEx_cheeger_diffNoiseLevels}
\end{figure}

\paragraph{\bf Impact of Observational Noise}
In this experiment, we examine the robustness of the loss function based on the De Giorgi-type energy-dissipation functional~\eqref{eq:discretizedEnergyloss} with $\alpha=0.5$ in the presence of observational noise. To simulate noisy measurements, we contaminate the training data by adding Gaussian perturbations to the particle trajectories generated from the simulations of solving SDE~\eqref{eq:sde}. Specifically, each particle position is corrupted according to
\begin{equation}
    \bm x \leftarrow \bm x + \eps\nonumber,
\end{equation}
where $\eps \sim \mathcal{N}(0, \sigma \mathrm{I}_n|\bm x|_{\rm max})$. The parameter $\sigma$ controls the noise intensity and is referred to as the noise level throughout this paper.

In addition to perturbation of the positions of the particles, the observational data $\bm v$ are further corrupted by another independent Gaussian noise with the same noise level, namely,
\begin{equation}
    \bm v \leftarrow \bm v + \eps\nonumber,
\end{equation}
where, with a slight abuse of notation, $\eps$ denotes a different Gaussian random variable $\eps \sim \mathcal{N}(0, \sigma \mathrm{I}_n|\bm v|_{\rm max})$. This setup reflects a more realistic scenario in which both state variables and their observations are affected by measurement noise.

The learning results are presented in Fig.~\ref{fig:2dEx_cheeger_diffNoiseLevels}. As the noise level $\sigma$ increases from $0.1$ to $0.3$, the learned potential remains stable and accurate, indicating that the loss function with $\alpha=0.5$ exhibits strong robustness with respect to observational noise. These results demonstrate that the proposed approach can effectively tolerate moderate levels of noise without causing significant degradation in performance.

\paragraph{\bf Further comparison on different choices of the parameter $\alpha$}
In previous discussions, we have explored the behavior of the loss function for different values of the parameter $\alpha$, namely $\alpha=0$, $\alpha=0.5$, and $\alpha=1.0$. When $\alpha=0$, the method suffers from a lack of gradient information of the unknown potential function, leading to reduced effectiveness. Conversely, for $\alpha=1.0$, the absence of velocity information results in a more severe non-convex optimization problem, which makes the method less robust to early-stage observational data. To further investigate this, we now consider a comparison of $\alpha=0.25$, $\alpha=0.5$, and $\alpha=0.75$, with the goal of roughly determining whether $\alpha=0.5$ is indeed optimal. In this case, we retain the majority of the settings from the previous examples. Specifically, we use $Q=5$ Gaussian initial distributions with a standard deviation of $0.5$, a time interval of $[0.3,0.8]$, a particle count of $N=10,000$, and a bandwidth of $h=0.05$, all with observational noise level $\sigma=0.1$. 

Fig.~\ref{fig:2dEx_diffAlpha} compares the learned potential using the loss function \eqref{eq:discretizedEnergyloss} for different values of the weighting parameter $\alpha=0.25, 0.5, 0.75$. Both $\alpha=0.5$ and $\alpha=0.25$ are better than $\alpha=0.75$. Moreover, when examining the contour details, $\alpha=0.5$ also shows some improvement over $\alpha=0.25$. Overall, $\alpha=0.5$ appears to give the best performance among the tested choices.

\begin{figure}[htbp]
\centering
\includegraphics[width=0.9\textwidth]{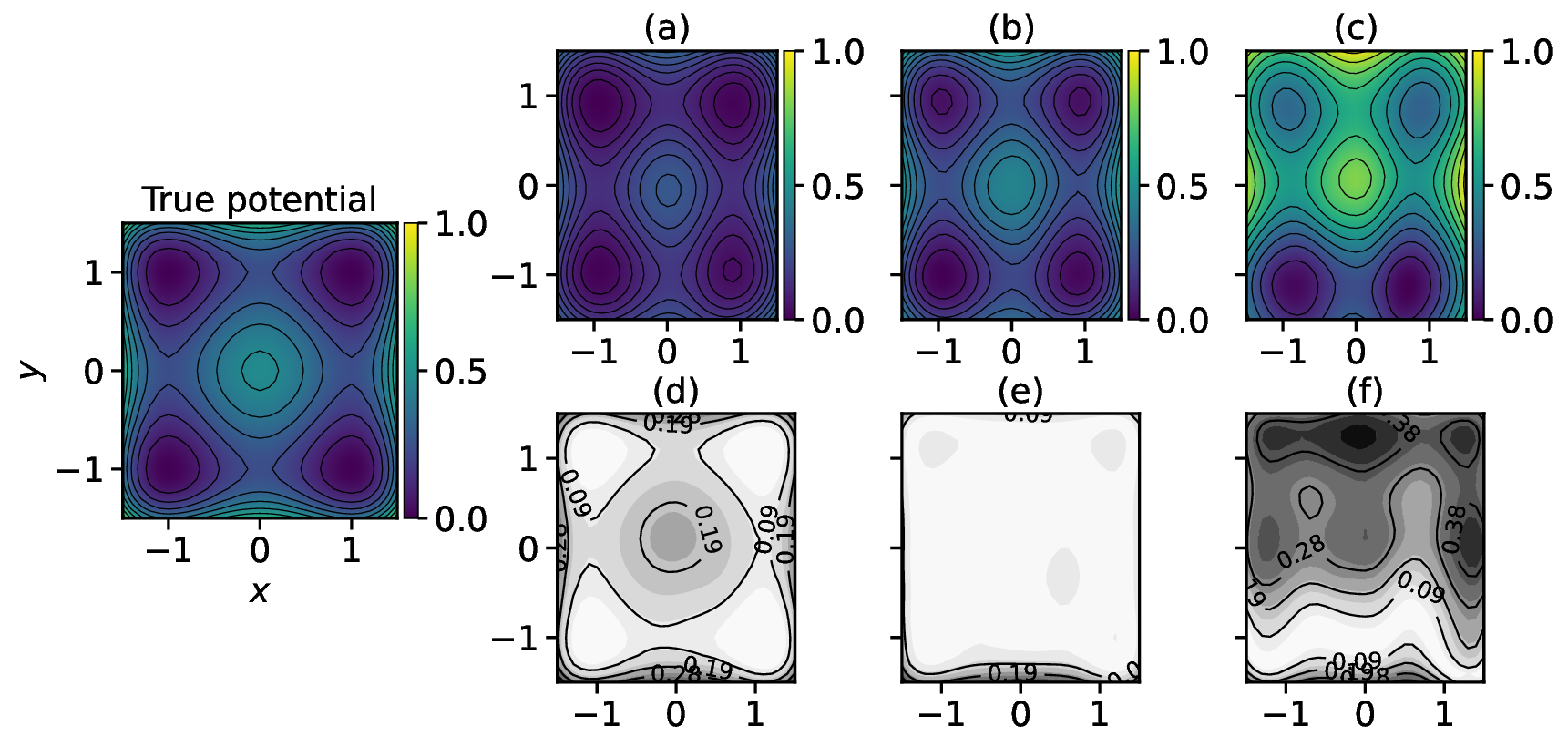}
\caption{\it A further comparison of the loss function \eqref{eq:discretizedEnergyloss} is made for different values of the weighting parameter $\alpha$. The leftmost panel shows the ground-truth potential. Panels (a)--(c) present the learned potentials obtained with $\alpha=0.25$, $\alpha=0.5$, and $\alpha=0.75$, respectively, while panels (d)--(f) show the corresponding pointwise absolute errors between the learned potentials and the ground-truth potential.}
\label{fig:2dEx_diffAlpha}
\end{figure}
\end{example}

\begin{example}
{\bf (Optimal Transport for estimating the velocity $\bm v$ from SDE Particle Snapshots)} The observations on velocity $\bm v$ in the above numerical experiments are computed via the force balance equation $\rho\bm v=-\rho\nabla\mu\approx-\rho[K_BT\nabla\ln(K_h*\rho_N)+\nabla\psi]$ while this is intractable in practice since the ground truth potential is not accessible. To address this issue, we present an example here to show that one can estimate the velocity $\bm v$ from the snapshots of the evolution of the SDE particles. first. Subsequently, this estimated velocity field can be used as input for our learning framework. It is worth mentioning that in our setting, we only have a limited number of snapshots. Specifically, just 6 snapshots over a fixed time period with a time step size of 0.1. Directly estimating the velocity field from such sparse data typically leads to suboptimal results, as the velocity field cannot be accurately reconstructed from such a small number of observations. This is especially true when we rely on a PDE loss directly based on the velocity field, as the estimation tends to be noisy and inaccurate. However, despite its roughness, the velocity field estimated from these sparse snapshots provides a surprisingly useful input to our method. 

In this example, the potential function \( V(x, y) \) is given by the sum of a radial term and a Gaussian mixture, as follows:
\begin{equation}\label{eq:2dGaussianMixtrue}
V(x, y) = \gamma \left( x^2 + y^2 \right) + \sum_{i=1}^{3} A_i \exp\left( -a_i \left( (x - x_i)^2 + (y - y_i)^2 \right) \right),
\end{equation}
where the parameters are chosen as:
\[
(c_1, c_2, c_3) = (0.2, 0.3, 0.3),
\]
\[
(A_1, A_2, A_3) = (-2.0, -2.0, -1.5),
\]
and the centers \( (\mu_{k,1}, \mu_{k,2}, \mu_{k,3}) \) are given by:
\[
(\mu_{k,1}, \mu_{k,2}, \mu_{k,3}) \in \{(-1, 0, 0), (1, 0, 0), (0, 1, 1)\},
\]
with the corresponding shape parameters \( (a_{k,1}, a_{k,2}, a_{k,3}) \) being:
\[
(a_{k,1}, a_{k,2}, a_{k,3}) \in \{(2, 1, 1), (2, 1, 1), (1, 2, 2)\}.
\]
The first term represents a radial potential centered at the origin, while the second term adds three localized Gaussian wells with the given amplitudes, centers, and shape parameters.

Similar to previous examples, we construct $Q=5$ Gaussian initial distributions with means uniformly sampled from the square $[-2,2]\times[-2,2]$ and a common standard deviation of $0.5$. For each initial distribution, we simulate the associated SDE~\eqref{eq:sde} using $N=2,000$ particles to generate sample trajectories. The system states are recorded at $M=6$ equally spaced time points over the interval $[0.3,0.8]$ with a fixed time step of $0.1$. The velocity data $\bm v$ are estimated via \emph{Optimal Transport} \cite{flamary2024pot,flamary2021pot} and the bandwidth $h$ of the kernel $K_h$ is still chosen as $0.05$. In addition, the observational noise level is set to $\sigma=0.1$.

Fig.~\ref{fig:2dEx_GaussianMixtrue_potential} compares the learned potential with the ground truth given by Eq.~\eqref{eq:2dGaussianMixtrue}. Fig.~\ref{fig:2dEx_GaussianMixtrue_traj} shows the evolution of particle trajectories within the learned potential. These trajectories were simulated using initial positions \(x_0\) and \(y_0\), which were drawn from normal distributions with a small standard deviation $0.1$. The particles' movements were governed by the stochastic differential equation given in Eq.~\eqref{eq:sde}, with the dynamics driven by the learned potential. Their positions were recorded at four distinct time points. Despite using only 2000 samples and employing the optimal transport method to estimate the velocity, our approach remains effective.
\begin{figure}[htbp]
    \centering
        \includegraphics[width=0.6\textwidth]{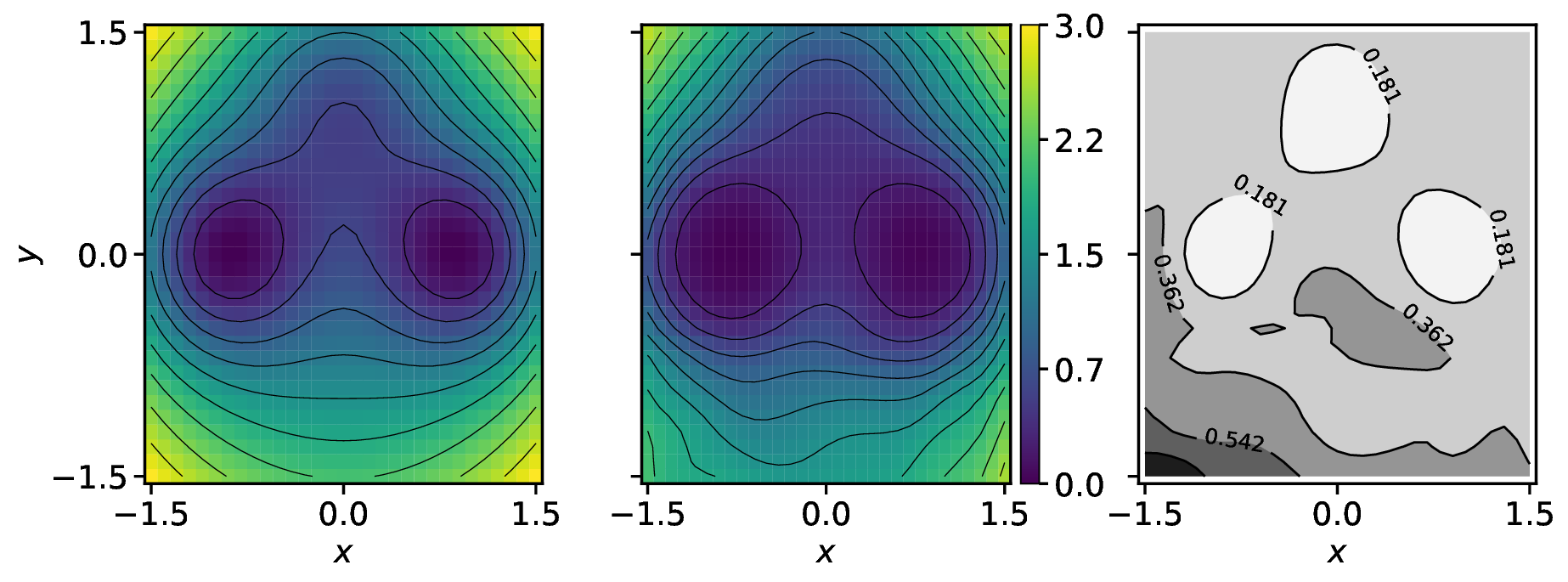}
    \caption{\it Comparison between the learned potential and the true potential given by Eq.~\eqref{eq:2dGaussianMixtrue}.}
    \label{fig:2dEx_GaussianMixtrue_potential}
\end{figure}

\begin{figure}[htbp]
    \centering
        \includegraphics[width=0.85\textwidth]{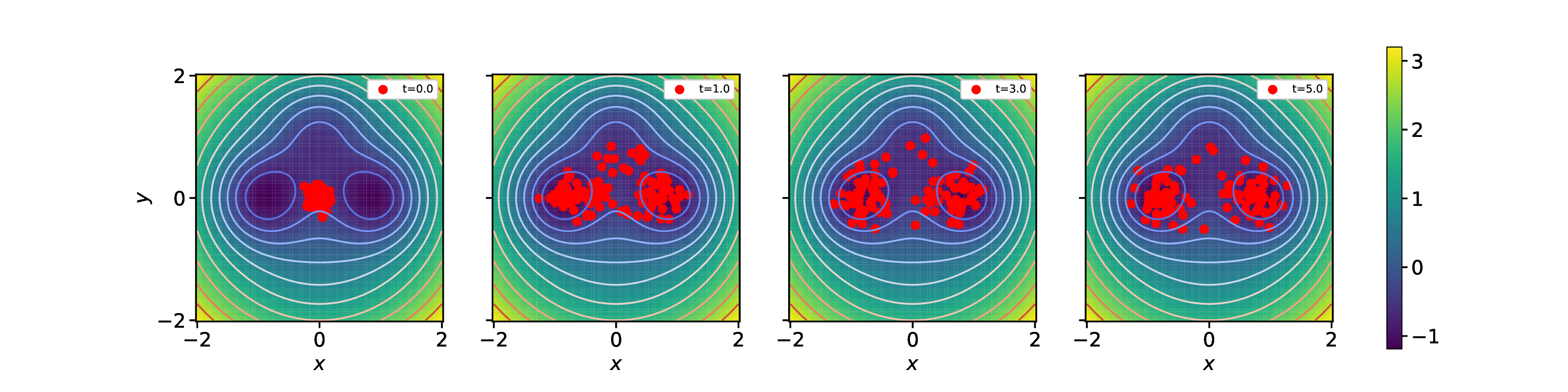}
    \caption{\it Evolution of particle trajectories within the learned two-dimensional potential. The red solid points represent the particle positions at four distinct time points, simulated using the stochastic differential equation given in Eq.~\eqref{eq:sde}, with the dynamics driven by the learned potential.}
    \label{fig:2dEx_GaussianMixtrue_traj}
\end{figure}
\end{example}

\begin{example}
    {\bf (Nonsymmetric Potential)} Following the symmetric double-well and quadruple-well examples, we further consider a nonsymmetric potential landscape to evaluate the capability of the proposed framework in recovering more general energy structures. Although the previous examples provide fundamental benchmarks, their symmetric configurations introduce equivalent metastable states and relatively simple transition patterns. To further assess the robustness of the proposed method, we consider a two-dimensional generalized diffusion system with an asymmetric multi-well potential given by
\begin{equation}
\begin{aligned}\label{eq:nonsymmetricPotential}
\psi(x,y)=&
-4\exp\left(-5((x+1)^2+(y+1)^2)\right)
-2.5\exp\left(-8(x^2+y^2)\right)\\
&-1.5\exp\left(-6((x-0.9)^2+(y-0.9)^2)\right)
+0.6x^2+0.9y^2\\
&+0.4xy+0.5x-0.3y .
\end{aligned}
\end{equation}

This potential contains three dominant wells with different depths and spatial locations, resulting in an intrinsically asymmetric energy landscape. Unlike the previous symmetric cases, the metastable states are no longer equivalent, and the associated diffusion dynamics involve biased transition pathways and non-uniform probability transport. Therefore, this example provides a more challenging test for the proposed learning framework, as it requires identifying both the global structure of the potential landscape and the local variations induced by the asymmetric interactions.

Similar to the previous examples, we construct $Q=5$ Gaussian initial distributions with means uniformly sampled from the square $[-2,2]\times[-2,2]$ and a common standard deviation of $0.5$. For each initial distribution, the corresponding SDE~\eqref{eq:sde} is simulated using $N=10,000$ particles to generate sample trajectories. The system states are recorded at $M=6$ equally spaced time points over the interval $[T_b,T_e]=[0.3,0.8]$ with a fixed time step of $0.1$. The velocity data $\bm v$ are obtained through the force balance relation, and the bandwidth parameter of the kernel $K_h$ is chosen as $h=0.05$, consistent with the previous experiments. 

Fig.~\ref{fig:2dEx_nonsymmetricPotential} presents the learned potential landscapes for the asymmetric potential under different noise levels, with $\sigma=0.0$, $0.1$, and $0.2$. It can be observed that the proposed framework recovers the main structures of the underlying potential across all noise levels. In particular, the learned potentials capture the locations and relative depths of the three metastable wells, as well as the overall asymmetric landscape of the target potential. 

As the noise intensity increases, small discrepancies can be observed in some local regions, especially near the transition regions between different wells, where the probability density exhibits more complicated transport behavior. Nevertheless, the dominant features of the potential landscape remain well preserved, demonstrating the robustness of the proposed approach against observational noise. These results indicate that the energy-based formulation provides an effective mechanism for extracting the intrinsic energetic structure of stochastic systems from noisy and finite observational data.

\begin{figure}[htbp]
\centering
\includegraphics[width=0.9\textwidth]{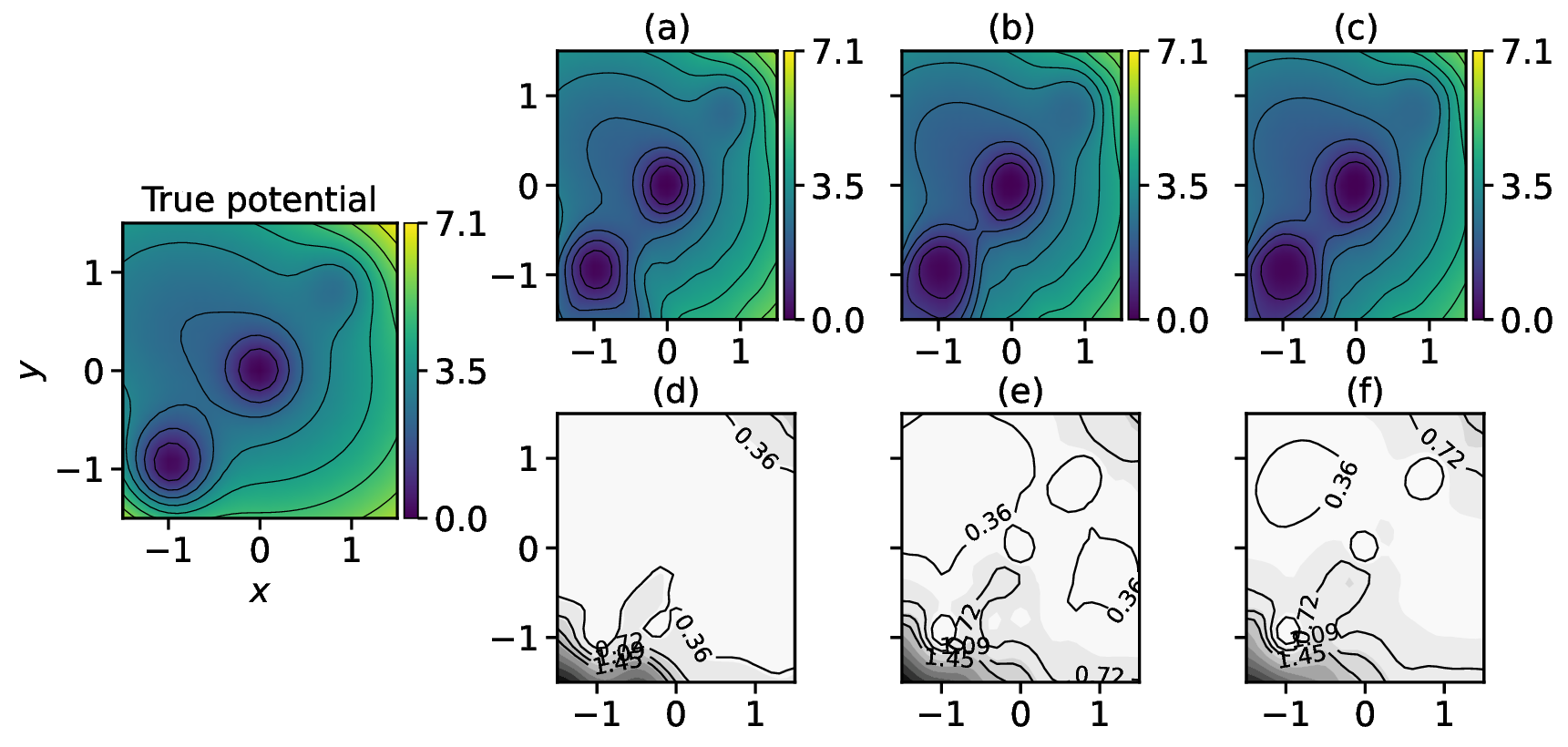}
\caption{\it Learning the nonsymmetric potential~\eqref{eq:nonsymmetricPotential} with $\alpha=0.5$ under different noise levels $\sigma$. The leftmost panel shows the ground-truth potential. Panels (a)--(c) present the learned potentials obtained with noise level $\sigma=0.0$, $0.1$, and $0.2$, respectively, while panels (d)--(f) show the corresponding pointwise absolute errors between the learned potentials and the ground-truth potential.}
\label{fig:2dEx_nonsymmetricPotential}
\end{figure}
\end{example}

\subsection{A 3D Example}
\begin{example}
In this example, we further conduct a more quantitative evaluation of the proposed method on a three-dimensional problem. The target potential function is defined as
\begin{equation}\label{eq:3d_potnetial}
    \psi(x,y,z)
    =
    \sum_{i=1}^3 c_i q_i^2
    +
    \sum_{k=1}^3 A_k \exp\!\left(
        -\sum_{j=1}^3 a_{k,j} (q_j - \mu_{k,j})^2
    \right),
\end{equation}
where $(q_1,q_2,q_3) = (x,y,z)$. The parameters are chosen as
$(c_1,c_2,c_3) = (0.2, 0.3, 0.3)$,
$(A_1,A_2,A_3) = (-2.0, -2.0, -1.5)$,
and
\[
(\mu_{k,1}, \mu_{k,2}, \mu_{k,3}) \in \{(-1,0,0), (1,0,0), (0,1,1)\},
\]
with the corresponding shape parameters
\[
(a_{k,1}, a_{k,2}, a_{k,3}) \in \{(2,1,1), (2,1,1), (1,2,2)\}.
\]
This potential consists of a quadratic confining term combined with multiple localized wells, resulting in a multimodal energy landscape in three dimensions.

To learn the target potential, we construct $Q = 10$ Gaussian initial distributions with means uniformly sampled from $[-2,2] \times [-2,2] \times [-2,2]$ and a common standard deviation of $0.1$. For each initial distribution, we simulate the associated stochastic differential equation~\eqref{eq:sde} using $N = 10{,}000$ particles to generate sample trajectories. The system states are recorded at $M = 6$ equally spaced time points over the interval $[0.3, 0.8]$ with a fixed time step of $0.1$.

The velocity data $\bm v$ are obtained using the force balance equation, and the bandwidth $h$ of the kernel $K_h$ is set to $0.05$, consistent with the previous experiments. The training data are contaminated with observational noise in the same manner as in the two-dimensional example described in the previous subsection, with the noise level set to $\sigma = 0.1$.

To visualize the three-dimensional potential, we present its values on three representative two-dimensional slices by fixing one coordinate at a time. Specifically, the slices are obtained by setting $z=0$, $y=0$, and $x=0$, respectively. In Fig.~\ref{fig:3dEx}, the first row shows the ground-truth potential evaluated on these slices, while the second row displays the corresponding potentials learned by our method with $\alpha=0.5$. As shown in Fig.~\ref{fig:3dEx}, the loss function~\eqref{eq:discretizedEnergyloss} with $\alpha=0.5$ remains effective in this three-dimensional setting, successfully capturing the underlying structure of the target potential despite the increased dimensionality and the presence of noise.

To further examine the performance of the proposed method, we re-simulate the stochastic differential equation \eqref{eq:sde} using both the true potential and the learned potential. In each case, we evolve $10{,}000$ particles and estimate the probability density by kernel density estimation with bandwidth $h=0.05$. A comparison of the resulting energy–dissipation rate $\frac{\dd E}{\dd t}(t)$ is shown in Fig.~\ref{fig:3dEx_energylaw}. The close agreement between the two curves indicates that the learned potential is able to reproduce the energy–dissipation behavior of the true system. In addition, as shown in \ref{fig:3dEx_energylaw_Wdist}, the Wasserstein distance between the learned and ground-truth particle distributions remain small and stable throughout the simulation. This suggests that the proposed variational learning framework successfully preserves the underlying energetic structure of the dynamics.

In addition to the energetic comparison, we quantitatively assess the accuracy of the learned potential by computing a weighted relative $L_2$ error between the gradients of the learned and true potential functions. Specifically, we define
\begin{equation}
\label{eq:weighted_L2_error}
\mathrm{Err}_{\nabla\psi}=
\frac{
\left(
\int_{\Omega}
\left|\nabla \psi_{\mathrm{learned}}(\bm x)-\nabla \psi_{\mathrm{true}}(\bm x)\right|^2\rho_{\mathrm{eq}}(\bm x)\dd \bm x
\right)^{1/2}
}{
\left(
\int_{\Omega}
\left|\nabla \psi_{\mathrm{true}}(\bm x)\right|^2\rho_{\mathrm{eq}}(\bm x)\dd \bm x
\right)^{1/2}
}\nonumber,
\end{equation}
where $\rho_{\mathrm{eq}}(\bm x) \propto e^{-\psi_{\mathrm{true}}(\bm x)}$ denotes the equilibrium distribution of the underlying stochastic gradient system. This choice of weighting emphasizes accuracy in regions of high probability mass that are most relevant to the long-time dynamics of the system.

Using this metric, the relative $L_2$ error of the learned potential gradient is $0.308$. Despite the presence of finite-sample effects and observational noise, this result demonstrates that the learned potential captures the dominant force structure of the system in a statistically meaningful sense, consistent with the observed agreement in energy–dissipation behavior.

While the above results demonstrate that the proposed variational framework is effective in capturing the dominant energetic and force structures of the system, achieving a more refined pointwise reconstruction of the potential function may benefit from incorporating additional information from PDE-based formulations, as well as from enriching the training data by increasing its diversity, for instance through a larger set of distinct initial distributions. In particular, loss functions derived directly from the governing Fokker–Planck equation can impose stronger local constraints on the dynamics when high-quality velocity or density data are available. Meanwhile, employing a richer collection of initial distributions (e.g., increasing the number of initial distributions $Q$) improves the exploration of the state space and can further enhance learning accuracy, consistent with the observations reported in Fig.~\ref{fig:2dEx_CP_diffICs}.

These observations suggest that the variational, energy-based approach and PDE-based learning strategies play complementary roles: the former offers robustness and structural consistency under partial or noisy observations, while the latter can enhance fine-scale accuracy when sufficient differential information is accessible. Developing principled hybrid methodologies that combine these two perspectives in a coherent manner represents a promising direction for future research, which we leave for further investigation.
\begin{figure}[htbp]
    \centering
        \includegraphics[width=0.75\textwidth]{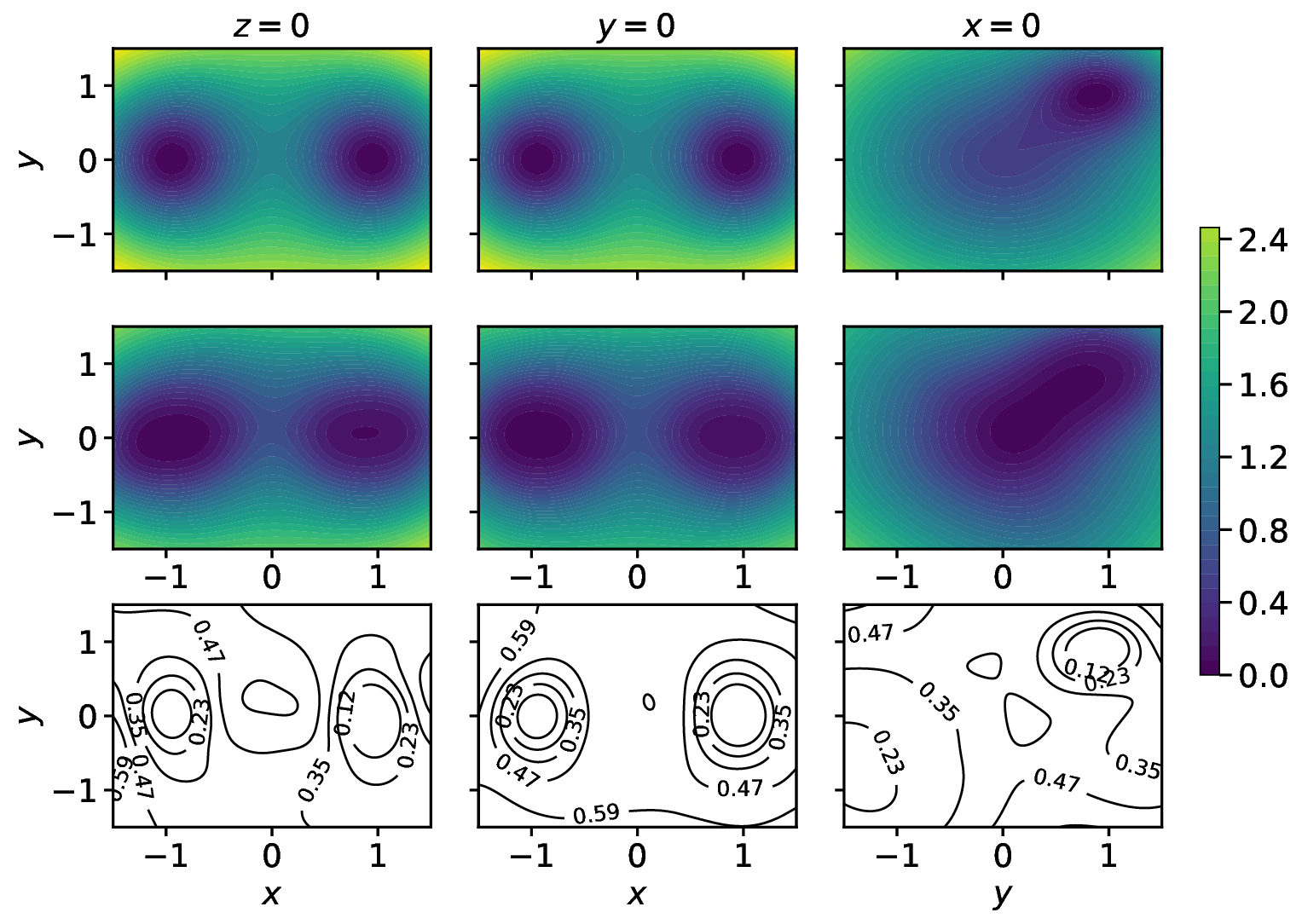}
    \caption{\it Learning the three-dimensional potential \eqref{eq:3d_potnetial} with observational noise level $\sigma = 0.1$. The three columns correspond to the density plot of the potential function in the $z=0$, $y=0$, and $x=0$ coordinate planes respectively. The first row shows the ground-truth potential, and the second row shows the learned potential. The third row shows the corresponding pointwise absolute errors between the learned potentials and the ground-truth potential.}
    \label{fig:3dEx}
    \end{figure}

 \begin{figure}
    \centering
    \begin{subfigure}[b]{0.47\textwidth}
        \centering
        \includegraphics[width=0.95\textwidth]{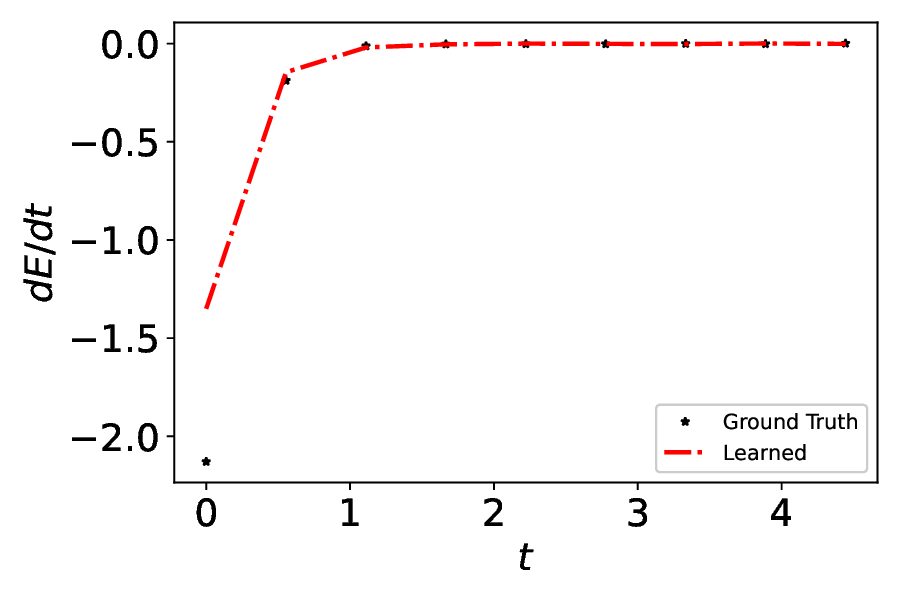}
        \caption{\it Comparison of the energy–dissipation rate $dE/dt$ associated with the learned and the ground-truth three-dimensional potentials \eqref{eq:3d_potnetial}.}
        \label{fig:3dEx_energylaw}
    \end{subfigure}
    \hfill
    \begin{subfigure}[b]{0.47\textwidth}
        \centering
        \includegraphics[width=0.95\textwidth]{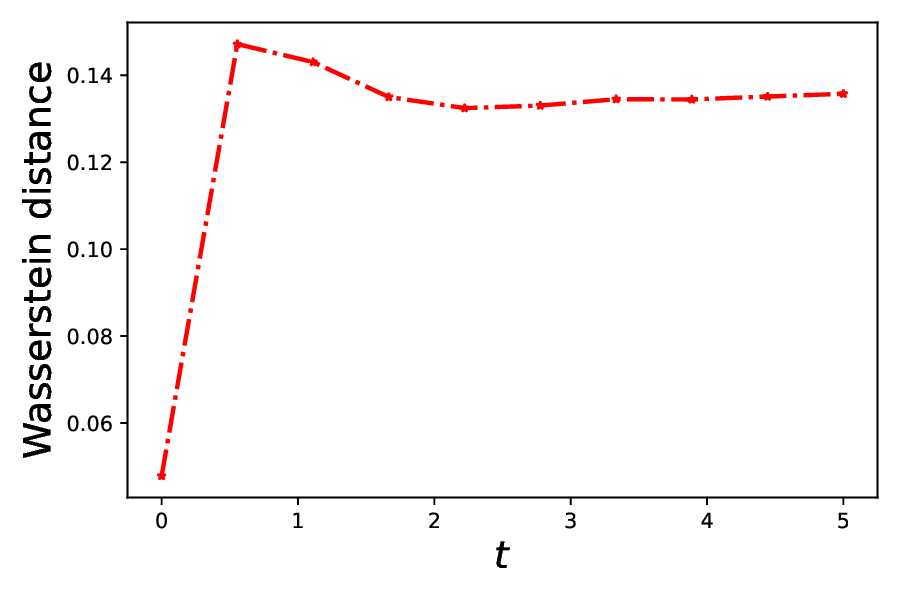}
        \caption{\it The Wasserstein distance between the learned particles and the ground-truth particles.}
    \label{fig:3dEx_Wdist}
    \end{subfigure}
    \caption{\it  Learning the three-dimensional potential \eqref{eq:3d_potnetial} with observational noise level $\sigma = 0.1$.}
\label{fig:3dEx_energylaw_Wdist}
\end{figure}
\end{example}

\subsection{A Simple Comparison with Direct Method}
As mentioned in Remark~\ref{rmk:direct}, when velocity observations $\bm v$ are available, one can apply the PDE-based loss function \eqref{eq:loworder_pde_loss} to learn the unknown potential function of a gradient system under random perturbations. However, this approach may not be suitable in all situations. As a proof of concept, we consider a simple scenario in this subsection.
\begin{example}
We revisit the one-dimensional example introduced in Section~\ref{sec:1d}, where the target potential is given by $\psi(x)=\frac{1}{4}x^4-\frac{1}{2}x^2$. We construct 5 Gaussian initial distributions ($Q=5$) with means uniformly sampled from the interval $[-2,2]$ and a common standard deviation of $0.5$. For each initial distribution, we simulate the associated stochastic differential equation~\eqref{eq:sde} using $N=10{,}000$ particles to generate sample trajectories. The system states are recorded at $M=6$ equally spaced time points over the interval $[0.3,0.8]$ with a fixed time separation of $0.1$.

The velocity data $v$ are computed according to $v = -\nabla \mu \approx -[\nabla \ln (K_h * \rho_N)+\nabla\psi]$, where this relation follows from the force balance equation $\rho v = - \rho \nabla \mu$. Here, the velocity is a scalar variable in 1D and $K_h * \rho_N$ denotes the kernel-smoothed empirical density with bandwidth $h = 0.05$. The resulting samples of the density $\rho$ and velocity $v$ collectively form the training dataset.

Furthermore, to model the influence of an external environment, the observational velocity data $v$ are additionally perturbed according to
\begin{equation}
v \leftarrow v + \lambda v_{\max} \sin(x),
\label{eq:velpert}
\end{equation}
where $v_{\max}$ denotes the maximum magnitude of the velocity $v$, and the parameter $\lambda = 0.8$ controls the amplitude of the imposed environmental velocity field.

Fig.~\ref{fig:1dEx_direct} presents the learning results obtained using the loss functions \eqref{eq:discretizedEnergyloss} with $\alpha=0.5$ and \eqref{eq:loworder_pde_loss}. It is evident that, under the presence of the environmental velocity perturbation~\eqref{eq:velpert}, the energy-based loss~\eqref{eq:discretizedEnergyloss} yields significantly more accurate reconstructions of the target potential than the PDE-based loss~\eqref{eq:loworder_pde_loss}.

This behavior can be explained by the structural difference between the two approaches. The PDE-based loss function relies directly on the observed velocity field and therefore cannot distinguish between the intrinsic velocity induced by the gradient of the chemical potential and the externally imposed environmental velocity field. As a result, the additional perturbation in the velocity observations leads to biased gradient information and degrades the learning performance.

In contrast, the energy-based loss function~\eqref{eq:discretizedEnergyloss} with $\alpha=0.5$ does not suffer from this limitation. The environmental velocity field does not enter the chemical potential term $\mu$ in the energy–dissipation formulation and only appears implicitly in the observational data. Consequently, the energy-based loss remains insensitive to such perturbations and exhibits improved robustness. This robustness is further reflected in its ability to tolerate noisy or contaminated velocity observations, leading to more reliable recovery of the underlying potential. 

This example highlights the advantage of energy-based formulations when the observed velocity field is subject to external or unmodeled influences.

 \begin{figure}[htbp]
    \centering
    \begin{subfigure}[b]{0.45\textwidth}
        \centering
        \includegraphics[width=0.9\textwidth]{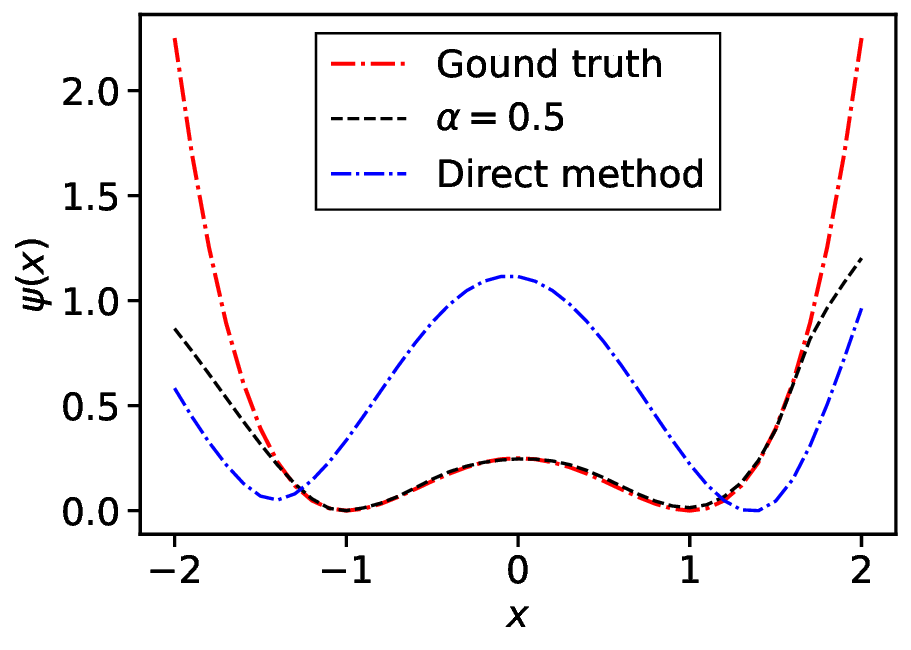}
        \caption{\it The learned potential.}
        \label{fig:1dEx_direct_potential}
    \end{subfigure}
    \hfill
    \begin{subfigure}[b]{0.45\textwidth}
        \centering
        \includegraphics[width=0.9\textwidth]{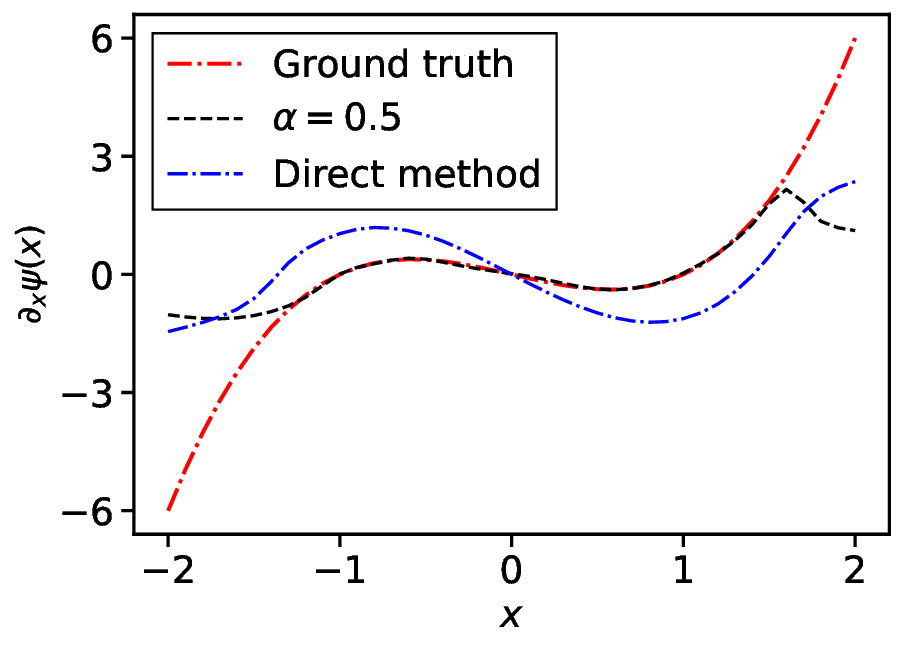}
        \caption{\it The learned gradient of the potential.}
    \label{fig:1dEx_direct_gradient}
    \end{subfigure}
    \caption{\it  Learning the potential $\psi(x)=\frac{1}{4}x^4-\frac{1}{2}x^2$ using the loss functions \eqref{eq:discretizedEnergyloss} with $\alpha=0.5$ and \eqref{eq:loworder_pde_loss}.}

\label{fig:1dEx_direct}
    \end{figure}

\end{example}

\section{Concluding Remarks}
\indent In this work, we have proposed a structure-aware learning framework for inferring unknown potential functions in stochastic gradient systems, based on the energy–dissipation law of generalized diffusion processes. The key idea is to explicitly distinguish and exploit different physical components of the dynamics, in particular the kinematic relations and the force-balance structure, within a variational formulation. This separation leads to a loss function that faithfully reflects the underlying energetic and dissipative mechanisms of the system.

Compared with our previous work \cite{lu2025EnVarA}, the present formulation yields a substantial improvement in both effectiveness and robustness when learning from finite, noisy, and partially corrupted data. In addition to numerical performance, our results emphasize the importance of disentangling distinct physical mechanisms in data-driven modeling. By respecting the intrinsic variational structure of the dynamics, the proposed approach enhances not only learning stability but also the interpretability of the inferred models.

The proposed construction is not restricted to overdamped Langevin dynamics.
More generally, the same idea can, in principle, be applied to generalized
gradient-flow systems characterized by an energy--dissipation law and
appropriate kinematic constraints. Once the free energy and dissipation
mechanism are specified, an analogous learning objective can be constructed
from the corresponding time-integrated energy--dissipation relation. This
provides a possible route for extending the present method to systems with
different free energies, mobility laws, or nonlinear dissipation mechanisms.
Nevertheless, the practical effectiveness of such extensions depends on the
observability of the relevant state and rate variables, the identifiability of
the unknown constitutive components, and the robustness of their discrete
approximations. A systematic investigation of these questions is left for
future work.

\section*{Acknowledgments}
Y.~Lu and X.~Li are partially supported by the Department of Energy Advanced Scientific Computing Research (ASCR) program under Award Number DE-SC0022276. C.~Liu is partially supported by NSF DMS-2216926 and DMS-2410742.
Q.~Tang is partially supported by the Department of Energy ASCR program under DOE-FOA-2493 ``Data-Intensive Scientific Machine Learning'' and Office of Science Early Career Research Program under Award Number DE-SC0026277. Y.~Wang is partially supported by NSF DMS-2153029 and DMS-2410740.

\section*{CRediT authorship contribution statement}
\textbf{Yubin Lu}: Conceptualization, Methodology, Software, Validation, Writing – original draft. \textbf{Xiaofan Li}: Conceptualization, Methodology, Writing – review and editing, Funding acquisition. \textbf{Chun Liu}: Conceptualization, Methodology, Writing – review and editing, Funding acquisition. \textbf{Qi Tang}: Conceptualization, Methodology, Writing – review and editing, Funding acquisition. \textbf{Yiwei Wang}: Conceptualization, Methodology, Writing – review and editing, Funding acquisition.

\section*{Declaration of generative AI and AI-assisted technologies in the writing process}
During the preparation of this work the author(s) used ChatGPT in order to improve the readability and language of the manuscript. After using this tool/service, the author(s) reviewed and edited the content as needed and take(s) full responsibility for the content of the published article.



\bibliographystyle{abbrv}
\bibliography{refs, VD} 
\setcounter{equation}{0}
\renewcommand{\theequation}{\Alph{section}.\arabic{equation}}
\newpage
\appendix
\onecolumn


\end{document}